%% file: main.tex
\documentclass[11pt]{article}

\usepackage[preprint]{acl}

\usepackage{times}
\usepackage{latexsym}

\usepackage[T1]{fontenc}

\usepackage[utf8]{inputenc}

\usepackage{microtype}

\usepackage{inconsolata}

\usepackage{graphicx}
\usepackage{booktabs}
\usepackage{amsmath}
\usepackage{multirow}
\usepackage{amssymb}
\usepackage{tcolorbox}
\usepackage{fancyvrb}
\usepackage{tabularx}
\usepackage{xcolor}
\usepackage{colortbl}
\newcommand{\cmark}{\checkmark}
\newcommand{\xmark}{$\times$}

\title{STRIVE: Probing Reasoning Limits in Graded Plausibility \\Generation and Evaluation
}

\author{
    \textbf{Bhiman Kumar Baghel\textsuperscript{1}} \quad
    \textbf{Anna Chrabaszcz\textsuperscript{1}} \quad
    \textbf{Tessa Warren\textsuperscript{1}} \quad
    \\
    \textbf{Michael Walsh Dickey\textsuperscript{1}} \quad
    \textbf{Haley C. Dresang\textsuperscript{2}} \quad
    \textbf{Xiang Lorraine Li\textsuperscript{1}}
    \\
    \textsuperscript{1}University of Pittsburgh,
    Pittsburgh, PA, USA\\
    \textsuperscript{2}University of Wisconsin--Madison,
    Madison, WI, USA\\
    \small{
   \textbf{Correspondence:} \href{mailto:bkb45@pitt.edu}{bkb45@pitt.edu}, \href{mailto:xiangli@pitt.edu}{xiangli@pitt.edu}
   }
}

\begin{document}
\maketitle
\input{latex/0_abstract}
\input{latex/1_introduction}

\input{latex/2_related_work}

\input{latex/3_framework}

\input{latex/4_experiments}

\input{latex/5_results}

\input{latex/6_analysis}
\input{latex/7_ablation}
\input{latex/8_conclusion}

\input{latex/9_limitations}
\input{latex/10_ethics}
\input{latex/11_AI_usage}

\bibliographystyle{acl_natbib}

\input{latex/main.bbl}
\appendix
\input{latex/12_appendix}

\end{document}

%% file: latex/0_abstract.tex
 
\begin{abstract}
Event knowledge concerns who does what to whom. Psycholinguists use event-plausibility judgments to examine how this knowledge supports human language processing. To isolate plausibility effects, these studies require controlled event sets in which one event slot varies across plausibility levels while all other event features remain fixed. Constructing such sets manually is labor-intensive. We therefore introduce STRIVE, an LLM-based framework for jointly generating and evaluating controlled event sets crossing plausibility class (plausible vs.\ implausible) with intended classification difficulty (easy vs.\ hard). Given a verb, STRIVE constructs a shared event frame, then produces one event per condition by varying one slot while holding all others fixed. In experiments with six models across 60 verbs, GPT-5.1 produced high-quality sets only 16.7\% of the time using the baseline generation prompt. Adding a global reasoning scratchpad and evaluator-guided refinement raised this rate to 75.0\%. Greater reasoning effort also improved evaluator--human agreement. Nevertheless, events near the plausibility boundary remain most difficult. They elicit the greatest human disagreement, and the best evaluator reaches only 57\% accuracy on the implausible-hard condition, indicating a need for human input. Overall, STRIVE offers a scalable approach to reducing manual effort by automating initial event-set generation and evaluation for psycholinguistic studies. \footnote{Under Review}
\end{abstract}

%% file: latex/1_introduction.tex
\input{figures/fig_conditions}

\section{Introduction}
\label{sec:intro}

Understanding and producing language relies on knowledge about real-world events, i.e., expectations about who does what to whom, with what, and where \citep{elman2019model}. Psycholinguistic studies probe event knowledge in humans by asking them to judge the plausibility of events described in sentences or depicted in images \citep{ivanova-etal-2021-eventsemantics,dresang2019semantic}. Here, event plausibility refers to the degree to which a complete described situation accords with event knowledge and ordinary world knowledge \citep{wang2018semantic,porada2021plausibility}. To attribute differences in these judgments to plausibility, such studies require matched verb in which only part of the event properties are modified. These sentences are refereed as stimuli. To capture finer distinctions in plausibility judgments, we construct gradient sentence stimuli using a two-by-two design crossing plausibility (plausible vs.\ implausible) with intended classification difficulty (easy vs.\ hard) (Figure~\ref{fig:conditions}). \footnote{These conditions are operational stimulus-design targets rather than natural semantic categories or calibrated probability intervals.}

To generate stimuli spanning four plausibility conditions for a single verb, we swap values of one of the event frame's slots. \footnote{We vary the agent in the main experiments and evaluate patient variation in \S\ref{sec:slot_dependency}.} This design localizes plausibility differences within a set to that slot and can support psycholinguistic research on lexical prediction and event integration \citep{mcrae2005basis,khalkhali2012integrating}, sentence comprehension \citep{bicknell2010effects,warren2015comprehending}, and language and cognitive impairments \citep{dresang2019semantic}. Figure~\ref{fig:conditions} shows the root verb \textit{catch} expanded into a shared soccer scene, with a different agent selected for each condition. Constructing such controlled, multilevel sets manually is labor-intensive, motivating NLP-based generation.

To achieve this goal, we developed \textbf{STRIVE}, a framework for generating and evaluating controlled event-plausibility stimulus sets. In a valid set, each sentence must match its intended plausibility condition, while all non-target slots remain fixed across sentences. STRIVE evaluates both requirements and uses feedback to revise sets that fail either.

To determine whether generation benefits from considering multiple agents, evaluator-guided refinement, or their combination, we compare four generation strategies. Reasoning Base (\textbf{RB}) generates a stimulus set in one structured pass. Reason-to-Verbalize (\textbf{R2V}) uses verbalized sampling \citep{zhang2026verbalized} to generate several fillers for the target slot before selecting one for each condition. Reason-to-Refine (\textbf{R2R}) revises an RB output using feedback from a separate evaluator \citep{madaan2023selfrefine,wang-etal-2025-cross}, while R2V-initialized R2R (\textbf{R2VR}) applies the same refinement to an R2V output. All four strategies use an expert-designed prompt that requires a global reasoning scratchpad and per-stimulus rationales \citep{wei2022chain,nye2022show}. For evaluation, we vary model family and reasoning effort and use the Alternative Annotator Test (AAT) \citep{calderon-etal-2025-alternative} to assess whether model-human agreement is non-inferior to human-human agreement.

Our experiments yield three findings. (F1) Generation: GPT-5.1 produces 28.3\% high-quality stimulus sets under RB and 75.0\% under R2R, a 2.6$\times$ improvement. Removing the global reasoning scratchpad reduces RB performance to 16.7\% and R2R performance to 40.0\%, while the full methods require 2.7--3.5$\times$ more output tokens (Appendix~\ref{app:token_costs}). (F2) Evaluation: Across six evaluator models, reasoning generally improves model-human agreement, but only GPT-5.1 and Sonnet~4.6 pass the primary AAT criterion, with bootstrap 5th percentiles above the $-0.05$ threshold. This indicates that reasoning effort alone is insufficient and that underlying model capabilities also matter. (F3) Boundary difficulty: cases nearest the plausibility boundary elicit the greatest human disagreement and remain the hardest for LLM evaluators, with the best evaluator reaching only 57\% accuracy on the implausible-hard condition (\S\ref{sec:eval_validation}).

To our knowledge, STRIVE is the first framework to jointly generate and evaluate matched, slot-controlled stimulus sets across four event-plausibility conditions. More broadly, STRIVE's root-verb-to-frame design provides a scalable approach to controlled stimulus construction across psycholinguistic studies of event knowledge.\footnote{The present experiments are restricted to English; cross-linguistic extension requires language-specific validation.}

%% file: figures/fig_conditions.tex
\begin{figure}[t]
\centering
\includegraphics[width=\columnwidth]{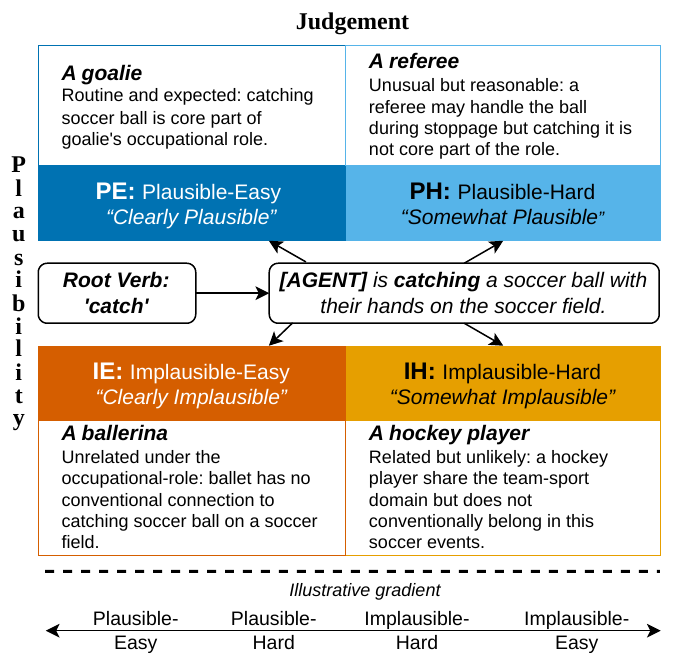}
\caption{STRIVE's four plausibility conditions for the root verb \textit{catch}. The verb form, patient, instrument, and location remain fixed while the agent varies. Easy and Hard indicate intended distance from the plausibility boundary; Clearly and Somewhat are annotator-facing labels. The gradient shows intended ordering, not calibrated probabilities or fixed semantic boundaries.}
\label{fig:conditions}
\end{figure}

%% file: latex/2_related_work.tex
\section{Related Work}
\label{sec:related}

This section focuses on work most directly related to plausibility event generation and evaluation. Appendix~\ref{app:detailed_related_work} discusses broader psycholinguistic and plausibility research.

\paragraph{Plausibility generation.}
ADEPT \citep{emami2021adept} pairs a sentence with a version formed by adding an adjective to a noun, then assigns the pair a five-way label indicating how the adjective affects plausibility. These labels encode changes within sentence pairs, not five matched conditions in a shared event frame. \citet{eichel-schulte-im-walde-2023-dataset} create pseudo-implausible events by replacing two constituents of corpus-derived event triples and collect absolute ratings. PRobELM \citep{yuan2024probelm} ranks alternatives constructed from Wikidata, while \citet{tang2023lesslikely} target one generation band of relevant but less-likely hypotheses. Unlike these methods, STRIVE generates all slot values from a root verb and constructs a matched four-condition set while holding non-target slots fixed and varying only the target role.

\paragraph{Plausibility evaluation and refinement.}
Existing methods estimate the plausibility of individual items using sentence probabilities \citep{kauf2024logprob}, direct LM judgments \citep{amouyal2024pretesting}, or a dedicated estimator such as VERA \citep{liu2023vera}. Self-Refine \citep{madaan2023selfrefine} improves generated outputs using self-feedback, while Cross-Refine \citep{wang-etal-2025-cross} uses a separate critic. STRIVE connects these lines by using feedback on both condition assignment and set-level role control to revise matched four-condition sets, while validating the evaluator against human judgments.

%% file: latex/3_framework.tex
\section{STRIVE Framework}
\label{sec:pipeline}

STRIVE has two parts: a \textit{generator} that produces graded plausibility stimuli from a root verb, and an \textit{evaluator} that serves two purposes: quality assessment of generated stimuli, and structured feedback that drives iterative refinement (§\ref{sec:refinement}), as illustrated in Figure~\ref{fig:methods}. Each is a standalone structured prompt\footnote{The prompts use \textit{thematic fit} as a heuristic for reasoning about agent substitutions within a fixed event frame. This heuristic supports candidate construction, but it is not the evaluated construct: all reported labels and human judgments assess complete-event plausibility.} designed through iterative human-AI collaboration between psycholinguists and AI researchers, refined based on failure patterns observed on three random verbs (\textit{catch}, \textit{break}, \textit{erase}). Although we instantiate the variable slot as the \textit{agent} throughout this paper, the framework generalises to other slots; we discuss patient-varying generation as a natural extension in §\ref{sec:slot_dependency}.

\input{figures/fig_methods}

\subsection{Generator}
\label{sec:generator}

Given a root verb $v$, the generator produces a set $\mathcal{S} = \{s_\text{PE},\, s_\text{PH},\, s_\text{IH},\, s_\text{IE}\}$ of four sentences that share a fixed five-slot template:
\begin{equation}
  s = \langle\, \textit{Agent},\ v',\ \textit{Patient},\ \textit{Instrument},\ \textit{Location}\,\rangle
  \label{eq:template}
\end{equation}
where $v'$ is a chosen inflected form of $v$. Only the \textit{Agent} slot varies across the four sentences; the other slots are held constant. This single-variable design localizes within-set plausibility differences to the agent substitution, eliminating lexical and syntactic confounds that would otherwise complicate experimental interpretation. Figure~\ref{fig:conditions} defines the four conditions.

\subsubsection{RB (Reasoning Base)}
\label{sec:rb}

RB instantiates the generator as a single forward pass through a structured prompt that decomposes stimulus generation into five sequential subtasks, each depending on the output of the previous. We adopted this decomposition after preliminary experiments showed that directly prompting for the full stimulus set repeatedly produced outputs violating multiple task requirements simultaneously, with no clear leverage for targeted correction.

The sub-tasks are: 
(1) \textbf{Scene design}: fix the static part of the stimulus (the \textit{scene}) by choosing the verb form, patient, instrument, and location together, conditioned on whether the resulting frame can support the full plausibility gradient; for \textit{catch} in Figure~\ref{fig:conditions}, the verb form becomes \textit{is catching}, yielding the scene \textit{``[Agent] is catching a soccer ball with their hands on the soccer field.''} 
(2) \textbf{Agent selection}: select one agent per plausibility condition, conditioned on the scene; in the running example, \textit{a goalie}, \textit{a
referee}, \textit{a hockey player}, and \textit{a ballerina} are picked for PE, PH, IH, and IE respectively. 
(3) \textbf{Visual Distinctiveness}: Some psycholinguistic event-knowledge studies use image-based stimuli \citep{ivanova-etal-2021-eventsemantics,dresang2019semantic}, so agents intended for downstream image generation should be visually distinguishable. STRIVE therefore applies a text-based pre-filter based on visible cues such as occupational attire (Figure~\ref{fig:conditions}). This heuristic does not validate actual images; image generation and image-grounded validation are outside the scope of this work. 
(4) \textbf{Composition}: insert each selected agent into the shared scene frame to compose the four sentences. (5) \textbf{Validation}: answer a 16-item YES/NO checklist covering every constraint from the previous steps; any \textit{no} answer
must be rectified before output (Figure~\ref{fig:v7-agent-part5}).

The output begins with a free-form scratchpad where the model reasons through these sub-tasks before committing to the structured outputs that follow. The structured output contains the scene, the four agents (each with a per-stimulus rationale justifying its gradient placement), and the four composed sentences. The full prompt and output schema are in Appendix~\ref{app:rb_prompt}.

\subsubsection{R2V (Reason-to-Verbalize)}
\label{sec:r2v}

R2V replaces RB's direct agent selection (Step 2) with verbalized sampling \cite{zhang2026verbalized}, because we hypothesize that explicit per-condition score ranges keep candidates within their intended plausibility grade, preventing the cross-grade contamination that arises when grade boundaries are left implicit.\footnote{We use these verbalized values only as heuristic plausibility scores for organizing candidate generation. They are neither token-level model probabilities nor calibrated estimates of real-world event likelihood or human responses, and we do not evaluate their numerical accuracy.} We partition the $[0,1]$ scoring scale under three principles: i) clear gradient ordering with no overlap, (ii) anchored endpoints at PE and IE that fix the gradient to the extremes of $[0,1]$, and (iii) non-zero buffers between adjacent conditions. The exact range boundaries are a design choice; any partition satisfying these principles would induce the same gradient structure. We adopt PE: 0.80--1.00, PH: 0.40--0.65, IH: 0.15--0.35, IE: 0.00--0.05, with buffer widths reflecting conceptual proximity between adjacent conditions: PE--PH (0.15) and IH--IE (0.10) separate ``clearly'' from ``somewhat'' judgments (annotator-facing labels, Figure~\ref{fig:conditions}), while the narrower PH--IH buffer (0.05) sits between two adjacent ``somewhat'' conditions across the plausibility cut. The model first generates a pool of candidate agents spanning the full plausibility range, then selects one candidate per plausibility condition from within the corresponding region. The full prompt and output schema are in Appendix~\ref{app:r2v_prompt}.

\subsubsection{Iterative Refinement (R2R, R2VR)}
\label{sec:refinement}

R2R and R2VR augment RB and R2V with iterative refinement using evaluator feedback. The refinement prompt already contains the same task definitions and validation rubric as RB. Its feedback field does not repeat this rubric or include the full evaluator output; instead, it contains a compact, instance-specific diagnosis with per-condition verdicts, evaluator-inferred conditions, brief issue descriptions, relevant overlap findings, and gradient and scene verdicts. Refinement therefore adds error localization rather than new task rules. These diagnoses drive two refiner behaviours. Agent-level failures concern a specific agent: it belongs in a different condition than assigned (e.g., a hockey player placed in PH instead of IH), or it is not visually identifiable, or both. The refiner replaces only the failing agents, preserving the scene and the passing agents. Scene-level failures concern the entire scene: the scene does not support the full gradient (e.g., a domain so narrow that no agent can plausibly fill IH). The refiner discards the output and selects a new scene. R2R and R2VR begin with RB and R2V outputs, respectively, then use a common refinement prompt with an instruction to rectify rather than regenerate and a feedback-history placeholder that accumulates prior compact diagnoses to prevent regression (Appendix~\ref{app:ir_prompt}).

\subsection{Evaluator}
\label{sec:evaluator}

The evaluator scores a generated stimulus set against task requirements and produces structured feedback. Given a verb and four candidate stimuli, it returns three verdicts, evaluating correctness of three major generation components: the agents, their visual distinctiveness, and the scene holding them on the gradient.

(1) \textbf{Per condition adversarial verdict} (per condition): the evaluator articulates the strongest argument that the agent belongs one level higher and one level lower on the gradient; the verdict is CORRECT only if both counter-arguments are clearly weaker than the assigned placement, otherwise INCORRECT. Specifically for IH, the evaluator audits the five overlap dimensions against PE (occupational domain, setting, tools, patient type, physical actions) and applies the negative test (removing the shared verb, does the association persist?); an IH agent with fewer than two overlap dimensions is reclassified as IE. We selected this two-of-five cutoff by inspecting seed examples during prompt development; it is a tunable operational choice rather than a universal semantic threshold.

(2) \textbf{Visual distinctiveness verdict}: each agent is rated IDENTIFIABLE, AMBIGUOUS, or UNIDENTIFIABLE based on whether its profession is recognisable from a photograph against a plain white background. Visual distinctness is then checked across all pairs of agents. The visual verdict is ALL\_DISTINCT (all agents IDENTIFIABLE and all pairs distinct), PROBLEMATIC (any agent UNIDENTIFIABLE, or three or more pairs indistinct), or MOSTLY\_DISTINCT otherwise.

(3) \textbf{Scene assessment}: after per-condition evaluation, the evaluator assesses the scene on instrument correctness, prototype clarity, imageability, and gradient supportability; if a condition is INCORRECT but a viable replacement agent exists, the verdict is AGENT\_FIXABLE, otherwise NEEDS\_REDESIGN. The full prompt and output schema are in Appendix~\ref{app:evaluator_prompt}.

\subsection{Why Graded Plausibility Is Hard}
\label{sec:why_hard}

Three challenges compound across the framework. The IH constraint sits at a semantic knife's edge: too much overlap with PE pushes the agent into PH; too little drops it to IE. Semantic distinctness does not imply visual distinctness: two professions that read as different on paper may still look the same in a photograph, making them unusable as paired stimuli. And all four conditions must be jointly valid within one shared scene: a domain that is too narrow collapses IH because every adjacent professional becomes plausible, invalidating the entire set. The evaluator must detect all three failure types and distinguish between those requiring agent replacement and those requiring full scene redesign.

%% file: figures/fig_methods.tex
\begin{figure*}[t]
  \centering
  \includegraphics[width=\textwidth]{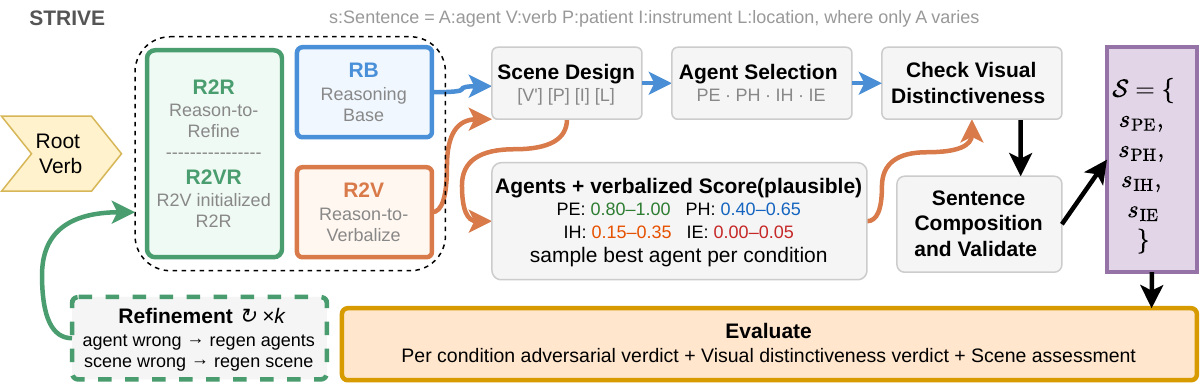}
  \caption{STRIVE framework. Given a root verb, the generator
produces a sentence set
$\mathcal{S} = \{s_{PE}, s_{PH}, s_{IH}, s_{IE}\}$ varying only
the agent across the four plausibility conditions. Four
generator variants share the pipeline (scene design, agent
selection, visual distinctiveness checking, composition):
\textbf{RB} is the single-shot reasoning base; \textbf{R2V}
adds verbalized plausibility scores over disjoint ranges per
condition and samples the best agent; \textbf{R2R} adds
iterative refinement that routes agent-level and scene-level
failures separately; \textbf{R2VR} initialises R2R with R2V.
The evaluator scores plausibility (adversarial check, IH overlap audit), visual evaluation, and scene assessment.}
  \label{fig:methods}
\end{figure*}

%% file: latex/4_experiments.tex
\section{Experiments}
\label{sec:experiments}

\subsection{Generator}
We run RB, R2V, R2R, and R2VR across six models on 60 picturable verbs drawn from three English verb-naming assessments \citep{choReyes2012verb,druks_masterson2000,swinburn2004comprehensive} and filtered to retain verbs that can instantiate all four event roles in our template (agent, patient, instrument, location). The assessments provide only the root verbs; STRIVE generates the verb form and all event-role values. The six models span scale and openness: GPT-5.1 and Claude Sonnet~4.6 (closed-source), Qwen3-Next-80B, Qwen3-30B, Ministral-3-14B, and Qwen3-4B (open-source). Iterative methods (R2R, R2VR) run up to $k=3$ iterations and exit early when the evaluator returns a GOOD scene verdict and all per-condition verdicts are CORRECT. We set temperature to 0.3 to balance two requirements: enough creativity to support diverse candidate selection in the middle of the gradient (PH and IH, where conceptual distinctions are subtler), while remaining deterministic enough for reliable results. For the same reason, native reasoning and thinking modes are disabled: these either constrain temperature to a fixed value, or, for closed-source models, return summarised traces rather than the verbatim reasoning. The scratchpad approach in §\ref{sec:generator} preserves full reasoning traces at our chosen temperature. Complete implementation details are in Appendix~\ref{app:hyperparameters}.

\subsection{Evaluation Experiments}
\label{sec:eval_experiments}

Evaluator outputs comprise per-condition, scene, and visual-distinctiveness verdicts, which do not yield a single comparable label for generation methods. We therefore map them to GOLD, SILVER, BRONZE, or FAIL using the hierarchical rule in Table~\ref{tab:tiers}, with plausibility as the prerequisite gate. Within STRIVE, GOLD denotes sets that pass all three evaluator checks. For downstream tasks, GOLD serves only as STRIVE's text-based filter; image-grounded and domain-expert validation remain required. We compare evaluator models and reasoning settings against human annotations.

\subsection{Human Annotation Study}
\label{sec:human_validation}

We collected human judgments mimicking the evaluator (§\ref{sec:evaluator}) on a subset of 30 stimulus sets (120 sentences) sampled across the GOLD, SILVER, and FAIL tiers from our main generation experiments. Eight undergraduate researchers in psycholinguistics\footnote{All annotators participated as volunteers; no compensation was provided.} each rated all 120 sentences for plausibility, yielding 960 sentence-level ratings, and assessed visual distinctiveness for each four-agent set. Both are categorical tasks, so we adopt unweighted Cohen's $\kappa$ \citep{Cohen1960ACO} as the primary metric for the AAT \citep{calderon-etal-2025-alternative}; this matches the tier-assignment success criterion (Table~\ref{tab:tiers}), which requires exact category match. Visual distinctiveness is decomposed into two sub-tasks. AgentID asks whether each agent's profession is identifiable from a photograph; the primary metric is 3-point unweighted $\kappa$. PairDist asks which agent pairs would look indistinguishable; the primary metric is binary $\kappa$, with Gwet's AC1 \citep{ac1} reported to address prevalence skew. Full survey details in Appendix~\ref{app:annotation}.

We use two evaluators to support both AAT-validated primary results and a fully open-source pipeline. Two configurations pass non-inferiority on the unweighted Cohen's $\kappa$ AAT against human annotators: GPT-5.1 (high reasoning effort) and Claude Sonnet~4.6 (non-thinking mode). We use GPT-5.1 as the primary evaluator for its lower per-token cost, applying its verdicts to all primary generation results in §\ref{sec:results}.
To validate that our generation findings do not depend on closed-source evaluation, we additionally evaluate every generator output with Qwen3-30B, the open-source evaluator whose unweighted Cohen's $\kappa$ with human annotators comes closest to the human-human $\kappa$ (Appendix~\ref{sec:evaluator_scaling}). Although this margin does not pass the AAT non-inferiority threshold, Qwen3-30B achieves a GOLD-tier concordance of 81.5\% with GPT-5.1 (Wilson 95\% CI excludes 50\%; details in Appendix~\ref{app:concordance}). This cross-evaluator agreement at the gold-stimulus decision level means the open-source pipeline recovers the same downstream design conclusions as the closed-source pipeline.
\input{tables/tab_tiers}

%% file: tables/tab_tiers.tex
\begin{table}[t]
\centering
\resizebox{\columnwidth}{!}{%
\begin{tabular}{llll}
\toprule
\textbf{Tier} & \textbf{Conditions} & \textbf{Scene} & \textbf{Visual} \\
\midrule
GOLD   & All 4 correct & \textsc{good}  & \textsc{all\_distinct}    \\
SILVER & All 4 correct & \textsc{good}  & \textsc{mostly\_distinct} \\
BRONZE & All 4 correct & Issues         & Any                       \\
FAIL   & $<$4 correct  & Any            & Any                       \\
\bottomrule
\end{tabular}}
\caption{Hierarchical quality-tier assignment. Plausibility is the first gate; scene and visual verdicts distinguish tiers only after all four conditions are correct.}
\label{tab:tiers}
\end{table}

%% file: latex/5_results.tex
\input{tables/tab_model_method}

\input{tables/tab_aat}
\input{figures/fig_rating_distribution}

\section{Results}
\label{sec:results}

\subsection{Generation Quality}
\label{sec:gen_results}

Table~\ref{tab:model_method} report GOLD rate of different models/methods judged by both GPT and Qwen and Table~\ref{tab:v7-agent-paper-examples} some samples. \textbf{Iteration is the central lever (F1).} Both judges agree that iterative refinement substantially improves GOLD rate over the standalone methods. Under the GPT judge, RB and R2V reach 11--13\% GOLD while R2R and R2VR reach 74--77\% on the two closed-source generators. Under the Qwen judge, standalone methods achieve 36--37\% GOLD across all six generators while iterative methods reach 65--68\%. The qualitative ranking RB $\approx$ R2V $\ll$ R2R $\approx$ R2VR holds across both evaluators.

\textbf{Refinement improves the plausibility gate.} We measure the fraction of sets with all four condition verdicts correct before applying scene and visual checks. Under the GPT judge, this rate rises from 51.7\% under RB to 86.7\% under R2R for GPT-5.1, and from 43.3\% to 89.1\% for Sonnet~4.6. Thus, refinement gains cannot be attributed only to the later checks. Appendix~\ref{app:plausibility_gate} reports all methods.

\textbf{Judge differences and caveats.} The two judges disagree on the magnitude of iteration's effect for the two generators where a clean cross-judge comparison exists. The Qwen judge is more generous than the GPT judge on standalone outputs from closed-source generators (51.7\% vs 28.3\% for RB) but less generous on their iterative outputs (48.7\% vs 74.8\% for R2R). For open-source generators on R2R/R2VR, GPT-judge evaluation was not run due to budget constraints. The iteration feedback in those cases also came from the Qwen judge, so the Qwen-judge column for those cells is a self-evaluation rather than an independent check; we therefore do not directly compare these GOLD rates against the closed-source rates. The Qwen judge serves as cross-evaluator trend confirmation on the closed-source pipeline, not as a substitute for the GPT judge.

\subsection{Evaluator Validation}
\label{sec:eval_validation}

\textbf{Humans distinguish the intended conditions, but IH remains most variable.} Before assessing evaluator--human agreement, we first ask whether human ratings differ overall across the four intended conditions. As expected, ratings become progressively less plausible from PE to IE, as shown by the observed means (PE 1.21, PH 1.89, IH 2.58, IE 3.86; Figure~\ref{fig:rating_dist}). A Friedman test \citep{friedman1937ranks} confirms that this overall difference is statistically reliable ($\chi^2(3)=76.5$, $p<10^{-15}$), showing that humans do not treat all four conditions alike. Because this overall effect could be driven only by the extreme conditions, we next ask whether humans distinguish each adjacent pair. Holm-corrected Wilcoxon signed-rank tests \citep{wilcoxon1945individual,holm1979simple} find reliable differences at every adjacent boundary (all adjusted $p<0.001$), including PH--IH. Thus, with 960 sentence-level ratings across 30 matched sets, the four conditions are both ordered and separable in human judgment, although IH remains the most variable (SD $=1.03$). Appendix~\ref{app:annotation} provides the test rationale and complete pairwise results.

\textbf{Plausibility classification: LLM and humans agree as much as humans agree with each other.} For the GPT-5.1 evaluator with high reasoning effort, L-H $\kappa$ matches the H-H baseline on the primary 4-point classification metric (0.530 vs 0.529, $\Delta = +0.001$; Table~\ref{tab:aat}). This passes the AAT non-inferiority test at $\delta \leq 0.05$: with 95\% confidence, the LLM is statistically no worse than a human annotator by more than 0.05 kappa. The supporting weighted $\kappa$ metric \citep{cohen1968weighted} passes at the tighter $\delta \leq 0.01$, confirming the LLM and humans agree on the gradient ordering: if humans rate a stimulus as PH, the LLM rates it PH or at worst PE, not IH or IE. The LLM evaluator can substitute for a human annotator at the primary classification task. Complete results is Appendix~\ref{app:aat_detail}.

\textbf{Visual distinctiveness validation is unreliable.}
Human-human agreement on visual distinctiveness is itself low ($\kappa = 0.277$ for AgentID, $0.407$ for PairDist), so humans do not form a reliable benchmark. On AgentID, the LLM tracks the human baseline closely but the baseline itself is weak. On PairDist, $\kappa$ shows a large LLM-human gap while Gwet's AC1 (which accounts for the high prevalence of distinguishable pairs) passes. We treat visual distinctiveness as unvalidated: the low human-human agreement reflects that annotators judged each profession's appearance from text alone, with each annotator imagining the visual details differently. Validation requires image-based annotation, presenting actual stimulus photographs to constrain this variability.

%% file: tables/tab_model_method.tex
\begin{table}[h]
\centering
\small
\resizebox{\columnwidth}{!}{%
\begin{tabular}{lcccc}
\toprule
\textbf{Model} & \textbf{RB} & \textbf{R2V} & \textbf{R2R} & \textbf{R2VR} \\
\midrule
GPT-5.1        & 28.3 / 51.7 & 13.3 / 56.7 & 75.0 / 41.7 & 71.7 / 60.0 \\
Sonnet 4.6     & 28.3 / 51.7 & 21.7 / 51.7 & 74.5 / 56.4 & 81.7 / 45.0 \\
Qwen3-80B &  3.3 / 31.7 &  6.7 / 33.3 &   — / 81.7  &   — / 66.7 \\
Qwen3-30B      &  6.7 / 41.7 & 10.0 / 31.7 &   — / 73.3  &   — / 83.3 \\
Ministral-14B  &  5.1 / 30.5 &  8.3 / 20.0 &   — / 70.0  &   — / 81.7 \\
Qwen3-4B       &  6.7 / 11.7 & 10.0 / 28.3 &   — / 66.7  &   — / 71.7 \\
\midrule
All models     & 13.1 / 36.5 & 11.7 / 36.9 & 74.8\textsuperscript{*} / 65.1 & 76.7\textsuperscript{*} / 68.1 \\
\bottomrule
\end{tabular}
}
\caption{GOLD (\%) per (model, method) cell, reported as
\emph{GPT judge / Qwen judge}. ``—'' indicates the GPT-judge
evaluation is unavailable due to budget constraints.
\textsuperscript{*}GPT-judge aggregate covers only the two
closed-source generators ($n{=}115$ for R2R, $n{=}120$ for R2VR)}
\label{tab:model_method}
\end{table}

%% file: tables/tab_aat.tex
\begin{table}[t]
\centering
\resizebox{\columnwidth}{!}{%
\begin{tabular}{ll rr r l}
\toprule
\textbf{Dimension} & \textbf{Metric} & \textbf{H-H} & \textbf{L-H} & \textbf{$\Delta$} & \textbf{Decision} \\
\midrule
\multicolumn{6}{l}{\textit{Plausibility (sentence-level, $n{=}120$)}} \\
\addlinespace[2pt]
4-pt (primary)  & $\kappa$   & 0.529 & 0.530 & $+$0.001 & \textbf{PASS} $\delta{\leq}0.05$ \\
4-pt (support)  & $\kappa_w$ & 0.806 & 0.821 & $+$0.015 & PASS $\delta{\leq}0.01$ \\
\midrule
\multicolumn{6}{l}{\textit{Visual distinctiveness}} \\
\addlinespace[2pt]
AgentID ($n{=}120$)  & $\kappa$   & 0.277 & 0.210 & $-$0.067 & PASS $\delta{\leq}0.15$ \\
PairDist ($n{=}180$) & $\kappa$   & 0.407 & 0.113 & $-$0.293 & \textbf{FAIL} \\
PairDist ($n{=}180$) & AC1        & 0.747 & 0.774 & $+$0.026 & PASS $\delta{\leq}0.01$ \\
\bottomrule
\end{tabular}}
\caption{AAT results. H-H = mean H-H pairwise agreement (28 pairs);
  L-H = mean LLM--human agreement (8 pairs);
  $\Delta = \bar{\kappa}_\text{L-H} - \bar{\kappa}_\text{H-H}$;
  Decision = smallest $\delta$ where bootstrap 5th percentile of
  $\Delta$ exceeds $-\delta$ ($B{=}10{,}000$).}
\label{tab:aat}
\end{table}

%% file: figures/fig_rating_distribution.tex
\begin{figure}[t]
\centering
\includegraphics[width=\columnwidth]{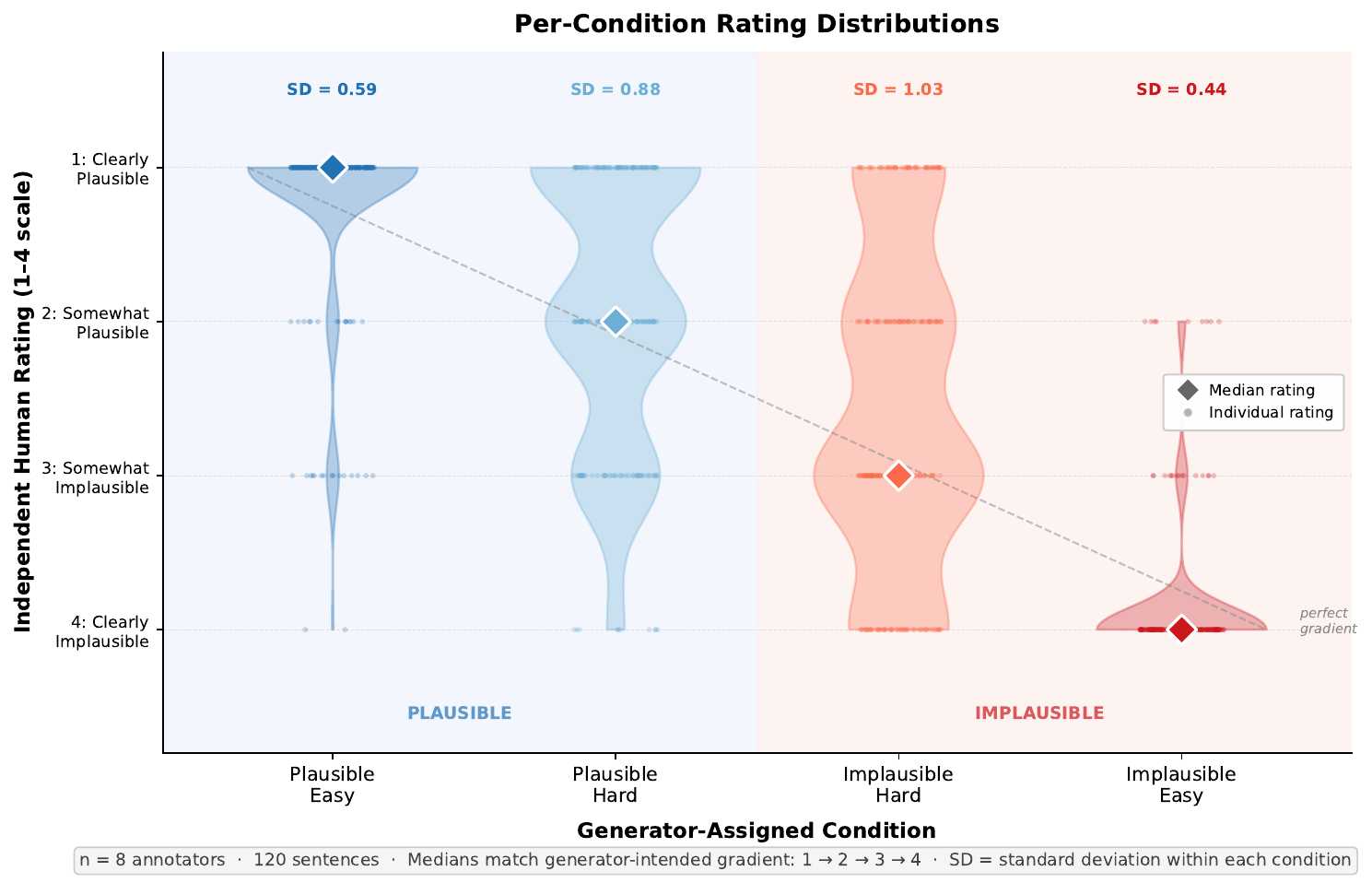}
\caption{Per-condition distribution of human plausibility ratings. Diamonds mark median
ratings; the dashed line traces the perfect gradient.
}
\label{fig:rating_dist}
\end{figure}

%% file: latex/6_analysis.tex
\section{Verb Difficulty Is Slot-Dependent}
\label{sec:slot_dependency}
Aggregating across all models and methods in agent-varying generation, per-verb GOLD rates span $42.4\%$ (\textit{throw}) to $3.4\%$ (\textit{tickle}, \textit{pay}), Table~\ref{tab:verb_difficulty}. The distribution divides into three bands: Easy ($\geq 20\%$ GOLD, $n=11$), Moderate (10--19\%, $n=34$), and Hard ($<10\%$, $n=15$). The Hard band concentrates two kinds of verbs: (a) actions tied to no specific occupation (\textit{eat}, \textit{drink}, \textit{tickle}, \textit{pay}), which leave no prototypical PE agent, and (b) actions with very narrow occupational scope (\textit{tow}, \textit{iron}, \textit{erase}), which leave no room for a distinct IH agent. Both failure modes are properties of the agent slot, not of the action itself.

A natural question arises from the per-verb GOLD distribution: does verb difficulty reflect an inherent property of the verb, or an artefact of fixing the agent slot? We therefore re-ran the pipeline with the patient slot varied. The framework is otherwise identical: only the five overlap dimensions and the visual identifiability criterion were recast from role-centered to object-centered properties. Across the same 60 verbs and five models (Sonnet 4.6 and the four open-source models), patient-varying R2V reaches $16.4\%$ GOLD versus $6.7\%$ for RB (Table~\ref{tab:slot_aggregate}). Crucially, only $3\%$ of model-verb pairs achieve GOLD in both slot conditions under R2V (Figure~\ref{fig:slot_perverb}): the agent-GOLD and patient-GOLD verb sets are largely disjoint. Verb difficulty is substantially slot-dependent, not verb-inherent. Agent-only generation has a structural ceiling set by which verbs admit a valid agent gradient; suggesting comprehensive stimulus coverage requires multi-slot variation.

%% file: latex/7_ablation.tex
\input{tables/tab_ablation}
\input{figures/fig_visual_pair}

\section{Ablations}
\label{sec:ablations}

We ablate two components of the generation framework using the GPT-judge evaluation: the visual identifiability constraint (\texttt{Viz}) and the reasoning scratchpad (\texttt{ResSP}). Results are in Table~\ref{tab:ablation}.

\textbf{The visual pre-filter strongly affects GOLD eligibility.} Under the text-based evaluator, removing the visual constraint from RB reduces GOLD\% to $\leq 1.7\%$ across all models. Without it, the model produces plausibility-appropriate agents predicted from text to look similar, e.g., a glazier and a home renovation contractor in generic workwear (Figure~\ref{fig:visual_pair}, left). Adding the constraint steers the model toward agents with more distinctive occupational cues (Figure~\ref{fig:visual_pair}, right): for GPT-5.1 on RB, GOLD\% rises from $0.0\%$ (no viz) to $16.7\%$ (viz, no scratchpad). This ablation measures compliance with the text-based pre-filter, not image-level distinctiveness, which requires image-grounded validation.

\textbf{Reasoning scratchpad consistently improves performance.}
The scratchpad lifts GOLD\% across models and methods. On single-shot methods, gains range from $+1.6$ to $+11.6$ across the open-source lineup (e.g., Qwen3-30B-A3B: $+3.4$ on RB, $+4.6$ on R2V; Ministral-14B: $+5.1$ on RB) and GPT-5.1 ($+11.6$ on RB, $+5.0$ on R2V). On iterative methods, GPT-5.1 gains $+35.0$ on R2R ($40.0 \to 75.0$) and $+41.7$ on R2VR ($30.0 \to 71.7$).\footnote{Open-source iterative cells are not in the ablation because those runs used Qwen-judge for iteration feedback (§\ref{sec:experiments}); we restrict ablation reads to GPT-judge for consistency.} The exception is Qwen3-Next-80B-A3B, which regresses by roughly $8$ points on both single-shot methods.

\input{tables/tab_slot_aggregate}

%% file: tables/tab_ablation.tex
\begin{table}[t]
\centering
\small
\setlength{\tabcolsep}{4pt}
\resizebox{\columnwidth}{!}{%
\begin{tabular}{l ccc cc}
\toprule
& \multicolumn{3}{c}{\textbf{RB}}
& \multicolumn{2}{c}{\textbf{R2V}} \\
\cmidrule(lr){2-4} \cmidrule(lr){5-6}
\textbf{Model}
& w/oViz & w/oResSP & wResSP
& w/oResSP & wResSP \\
\midrule
gpt-5.1         & 0.0 & \textbf{16.7} & \textbf{28.3} & 8.3           & 13.3          \\
Qwen3-Next-80B  & 0.0 & 11.7          & 3.3           & \textbf{15.0} & 6.7           \\
Qwen3-30B       & 1.7 & 3.3           & 6.7           & 5.4           & \textbf{10.0} \\
Ministral-14B   & 0.0 & 0.0           & 5.1           & 6.7           & 8.3           \\
Qwen3-4B        & 1.7 & 5.1           & 6.7           & 6.9           & \textbf{10.0} \\
\bottomrule
\end{tabular}}
\caption{GOLD\% by model $\times$ strategy on 60 verbs. \texttt{ResSP} = reasoning scratchpad; \texttt{Viz} =
visual identifiability constraint. \textbf{Bold} - each column's
best configuration.}
\label{tab:ablation}
\end{table}

%% file: figures/fig_visual_pair.tex
\begin{figure}[t]
  \centering
  \includegraphics[width=0.23\columnwidth]{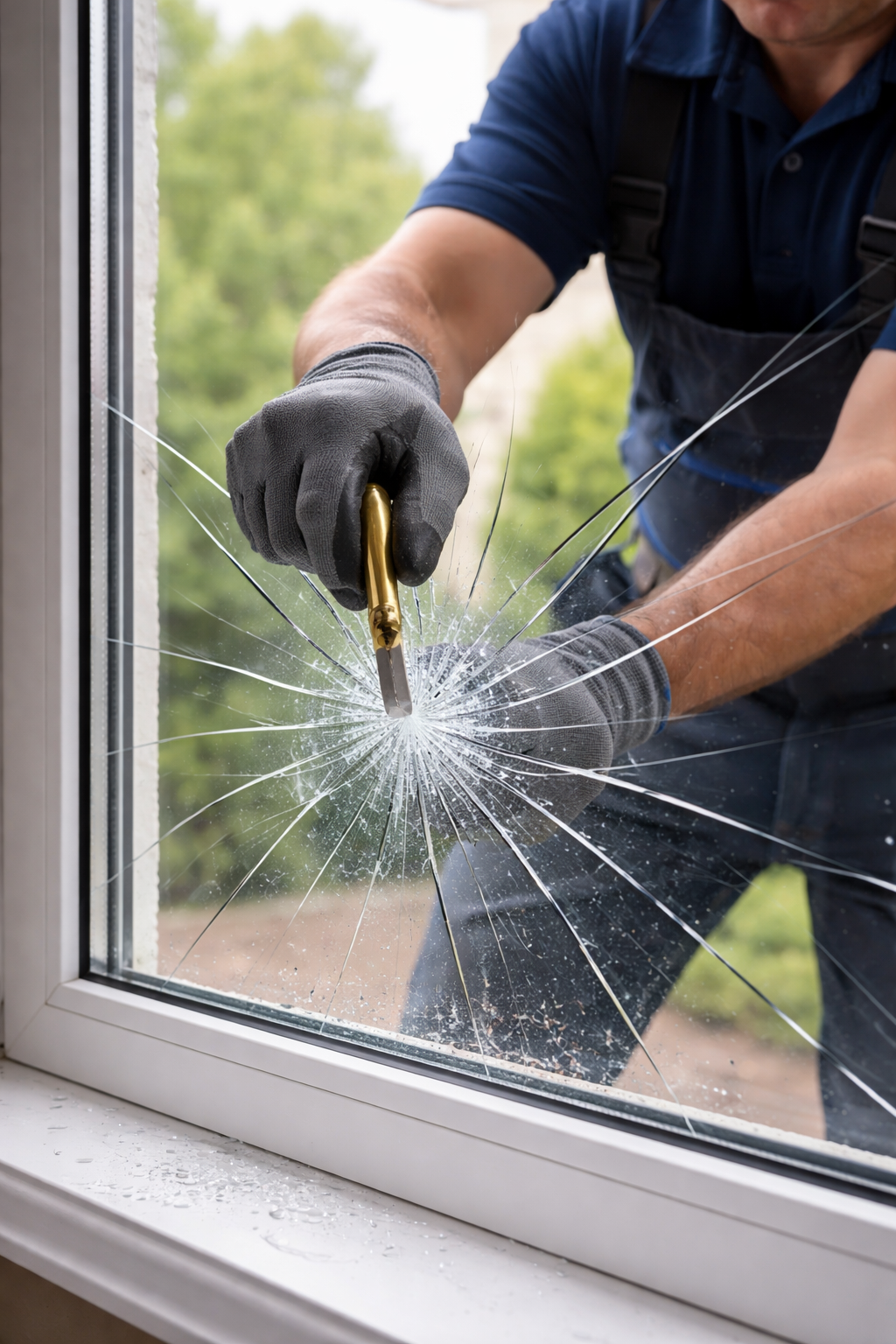}
  \hfill
  \includegraphics[width=0.23\columnwidth]{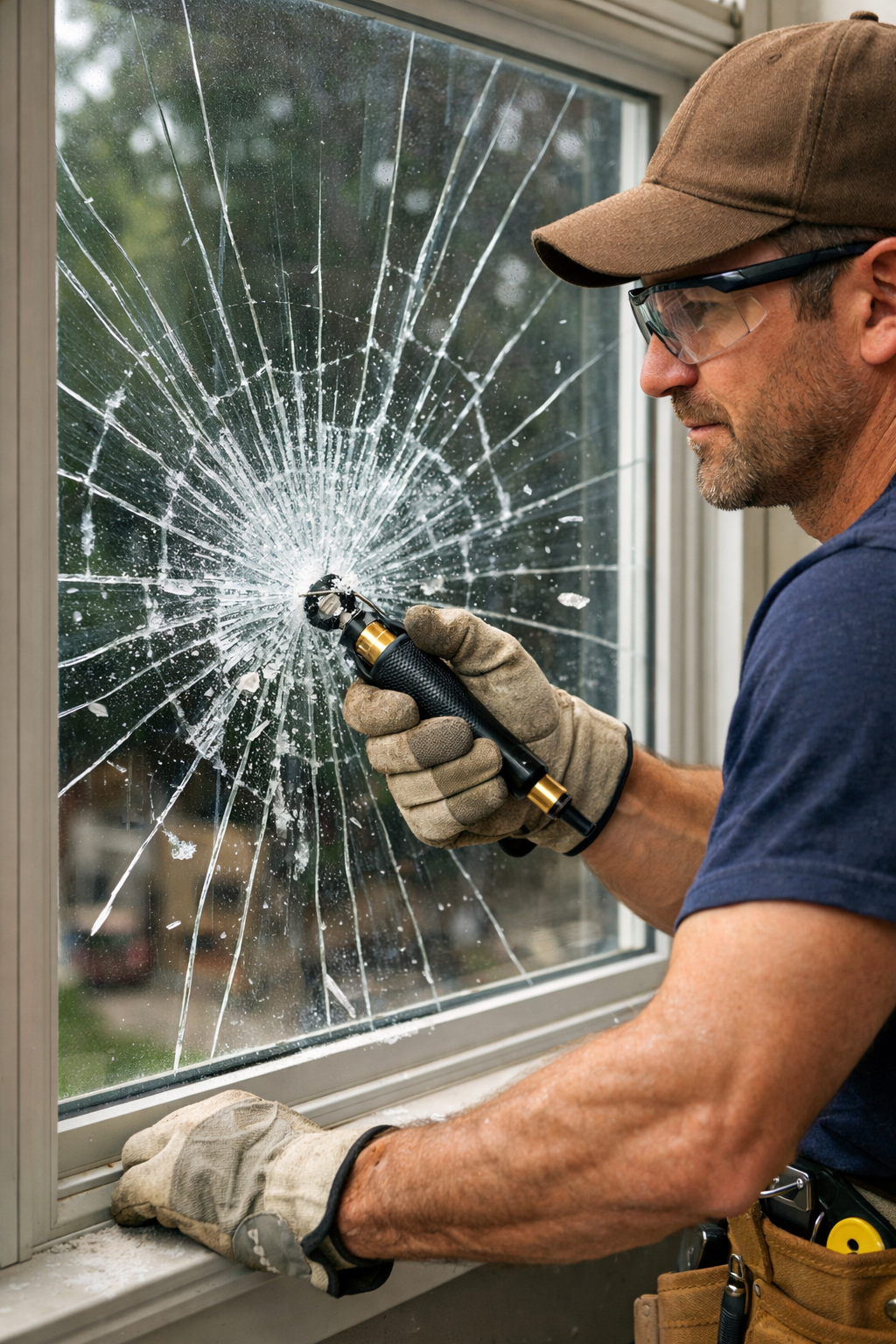}
  \hfill
  \includegraphics[width=0.23\columnwidth]{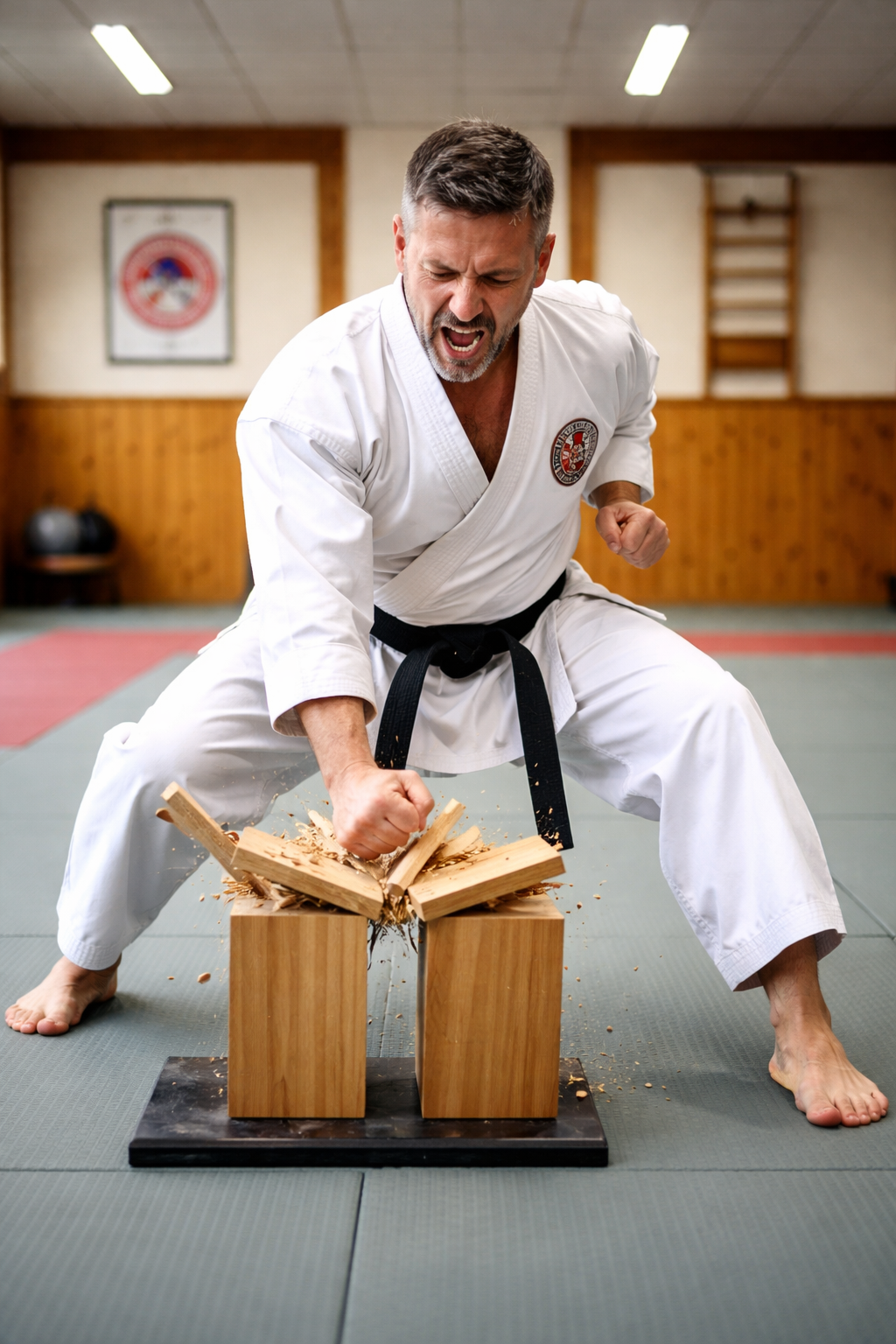}
  \hfill
  \includegraphics[width=0.23\columnwidth]{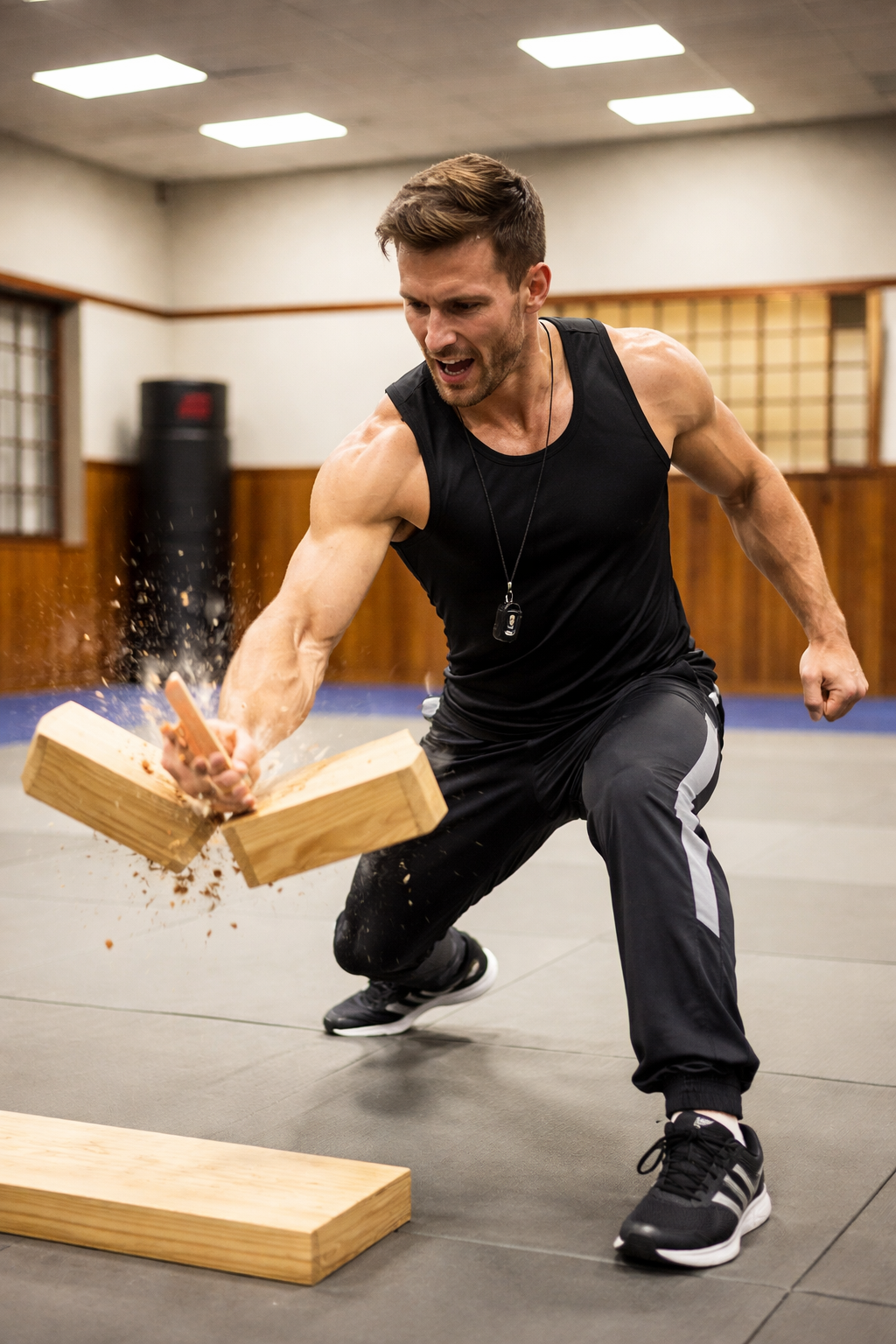}
  \caption{Visual distinctiveness failure and recovery for the
verb \textit{break}, visualized for illustration (STRIVE outputs
sentences, not images). \textbf{Left} (without visual
constraints): a glazier and a home renovation contractor,
satisfying their intended plausibility conditions but visually indistinguishable in
generic workwear. \textbf{Right} (with visual constraints):
a karate instructor and a personal trainer, satisfying
the same plausibility conditions but visually distinct through attire.}
  \label{fig:visual_pair}
\end{figure}

%% file: tables/tab_slot_aggregate.tex
\begin{table}[t]
\centering
\resizebox{\columnwidth}{!}{%
\begin{tabular}{l rr rr}
\toprule
& \multicolumn{2}{c}{\textbf{RB}} & \multicolumn{2}{c}{\textbf{R2V}} \\
\cmidrule(lr){2-3} \cmidrule(lr){4-5}
\textbf{Model} & \textbf{Agent} & \textbf{Patient} & \textbf{Agent} & \textbf{Patient} \\
\midrule
Sonnet 4.6         & 28.3 & 6.7  & 21.7 & 23.3 \\
Qwen3-Next-80B-A3B & 3.3  & 5.0  & 6.7  & 21.7 \\
Qwen3-30B-A3B      & 6.7  & 10.0 & 10.0 & 11.7 \\
Ministral-14B      & 5.1  & 5.2  & 8.3  & 11.9 \\
Qwen3-4B           & 6.7  & 6.7  & 10.0 & 13.3 \\
\midrule
\textbf{Mean}      & \textbf{10.0} & \textbf{6.7} & \textbf{11.3} & \textbf{16.4} \\
\bottomrule
\end{tabular}
}
\caption{Aggregate GOLD\% by slot condition and method, across
the five models with both-slot coverage. $n=60$ verbs per cell.}
\label{tab:slot_aggregate}
\end{table}

%% file: latex/8_conclusion.tex
\section{Conclusion}
\label{sec:conclusion}

We introduced STRIVE, a framework for jointly generating and evaluating matched four-condition event-plausibility stimulus sets under shared-frame and single-slot constraints. Our results show that structured evaluator feedback and explicit reasoning improve generation, while reasoning effort improves alignment between LLM and human plausibility judgments. These gains are not uniform: evaluation remains most difficult near the plausibility boundary, and generation difficulty depends on which event slot is varied. In particular, verbs that support successful agent-varying sets do not necessarily support patient-varying sets, showing that difficulty is not inherent to the verb. These findings position controlled graded-plausibility generation as a joint problem of event frame construction, slot choice, and fine-grained evaluation. STRIVE provides a scalable foundation for psycholinguistic stimulus design and motivates future methods that choose whether to vary the agent, patient, or another event slot for each verb while accounting for uncertainty near the plausibility boundary.

%% file: latex/9_limitations.tex
\section{Limitations}
\label{sec:limitations}

\paragraph{Text-based visual evaluation.}
The LLM evaluator and human annotators both judged visual
distinctiveness from text descriptions of agent attire rather
than from actual images. This is a deliberate design choice:
STRIVE produces sentence-level stimuli meant to be consumed by
any downstream image-generation model, so the visual
constraint serves only as a text-based pre-filter for downstream
image generation, not as validation of actual visual distinctiveness. The
low human--human agreement on the visual sub-tasks (AgentID
$\kappa = 0.277$, PairDist $\kappa = 0.407$,
§\ref{sec:eval_validation}) suggests that text descriptions leave
room for annotator imagination to diverge; image-grounded
validation on the downstream stimulus images is a natural
complement.

\paragraph{Occupational stereotyping.}
The visual-identifiability constraint steers agent selection
toward professions with standardised attire, which encodes
demographic and gender stereotypes (§\ref{sec:ethics}).
Diversifying the agent pool beyond such defaults remains an open
methodological question that interacts directly with the visual
constraint.

\paragraph{English-only scope.}
All experiments use English verbs and sentences because
the source verb-naming assessments are English-language
instruments. Extending STRIVE to other languages requires
language-specific validation of its event schemas and plausibility
conditions.

\paragraph{Clinical validation.}
STRIVE produces sentence-level stimuli intended for downstream
image generation and patient-facing assessment. Image generation,
behavioural validation with persons with aphasia, and clinical
reliability studies remain necessary before clinical deployment.

%% file: latex/10_ethics.tex
\section{Ethical Considerations}
\label{sec:ethics}

\paragraph{Clinical scope.}
STRIVE is designed to support psycholinguistic research and the
development of assessment materials for aphasia; it is not intended
for clinical diagnosis or autonomous clinical decision-making.
Any deployment involving patient populations must be conducted under
appropriate institutional ethical oversight and clinician supervision.
Generated stimuli should be reviewed by a qualified clinician before
use in patient-facing settings.

\paragraph{Occupational stereotyping.}
Agent selection in STRIVE relies on occupational roles with
distinctive visual presentations.
This design choice encodes stereotyped associations between
professions and appearance --- for example, defaulting to
gender-specific visual representations of certain roles.
These defaults may not reflect the diversity of actual practitioners
and could skew stimuli in ways that affect responses from patients
with different cultural or demographic backgrounds.
We flag this as a limitation requiring human review before clinical
deployment, and encourage future work to diversify the agent pool
beyond role stereotypes.

\paragraph{Annotator welfare.}
The human annotation study involved rating event plausibility from
text descriptions only; no sensitive, distressing, or harmful content
was included in the stimuli.
Annotators participated voluntarily,
and were not exposed to patient data or clinical material.
No personally identifiable information was collected.

\paragraph{Model and data transparency.}
All models used in this work are commercially available or
publicly released systems; no proprietary patient data were used at
any stage of the pipeline.
The generated stimuli, evaluation outputs, and annotation data will
be released for research reproducibility, subject to review to
ensure no harmful content is included in the release.

%% file: latex/11_AI_usage.tex
\section{Use of AI Assistants}

We used Claude (Anthropic) as a research, coding, and writing
assistant during this project. Claude assisted with brainstorming
and iterative design of the generator and evaluator prompts
(§\ref{sec:pipeline}), implementation of the STRIVE framework and
analysis code, and drafting and polishing prose in this
manuscript. All design decisions, scientific claims, and
analytical conclusions are the authors' own. The language models
studied as part of STRIVE appear in the paper as research
subjects, not as authorial assistants.

%% file: latex/12_appendix.tex
\input{appendix/J_detailed_related_work}
\input{appendix/A_rb_gen}
\input{appendix/B_r2v_gen}

\input{appendix/C_evaluator}

\input{appendix/D_experiment-deatils}
\input{appendix/E_evaluator-scaling}

\input{appendix/F_log-prob}
\input{appendix/G_verb-difficulty}
\input{appendix/H_token-cost}
\input{appendix/I_concordance}
\input{appendix/K_plausibility_gate}

\input{appendix/D_appendix_annotation}
\input{appendix/E_appendix_aat}

\input{appendix/v7_agent_paper_examples_companion}

%% file: appendix/J_detailed_related_work.tex
\section{Detailed Related Work}
\label{app:detailed_related_work}

\paragraph{Psycholinguistic uses of controlled event stimuli.}
Controlled event stimuli support several psycholinguistic paradigms. \citet{mcrae2005basis} show that nouns generate expectancies for verbs, and \citet{khalkhali2012integrating} examine the integration of words referring to typical event sequences. \citet{bicknell2010effects} study how event knowledge affects the processing of verbal arguments, while \citet{warren2015comprehending} examine the comprehension of impossible events. Across modalities, \citet{ivanova-etal-2021-eventsemantics} compare plausibility judgments for sentences and line drawings, and \citet{dresang2019semantic} develop an assessment of event-related conceptual semantic knowledge. These studies demonstrate the breadth of psycholinguistic uses for controlled event stimuli, but do not address their automated construction as matched, multilevel sets.

\paragraph{Relation to selectional preference and thematic fit.}
\textit{Selectional preference} describes a predicate's probabilistic expectations about the semantic properties of fillers in its argument slots \citep{resnik1996selectional}, whereas \textit{thematic fit} describes the graded compatibility of a particular filler with an event slot in context \citep{mcrae1998,pado-etal-2006-modelling,alshemali2024thematicfit}. Both constructs are broadly related to STRIVE's prompt design. Requiring an agent-slot filler to be a living human identified by a profession constrains the semantic class of possible fillers. In the worked \textit{catch} example, the model considers how well a goalie, referee, hockey player, or ballerina fills the agent slot in the fixed soccer-ball scene; this resembles thematic-fit reasoning. However, these choices are made internally by the model. STRIVE does not independently manipulate or score either construct, nor does it test how accurately the model represents them. Our evaluation concerns only the resulting complete-event plausibility, defined as the degree to which the complete situation accords with event knowledge and ordinary world knowledge. Selectional preference and thematic fit therefore describe possible mechanisms informing generation, not controlled variables or measured outcomes in this study.

\paragraph{Plausibility modeling and measurement.}
\citet{wang2018semantic} construct a 3,062-item dataset of subject, verb, and object triples from crowdsourced subject-verb and verb-object pairs and collect binary physical-plausibility labels. \citet{porada2019gorilla} learn plausibility from naturally occurring corpus events using self-supervision. \citet{porada2021plausibility} train on Wikipedia events paired with random argument perturbations and improve consistency across conceptual abstractions. These methods estimate the plausibility of individual events rather than construct graded matched sets. Graded resources operationalize different quantities. ADEPT \citep{emami2021adept} pairs a sentence with a modified version formed by adding an adjective to a noun, then assigns the pair a five-way label indicating how the adjective affects the event's plausibility. These labels encode changes within sentence pairs, not five matched conditions in a shared event frame. \citet{eichel-schulte-im-walde-2023-dataset} collect absolute slider ratings for corpus-derived original and pseudo-implausible event triples; because the midpoint cannot be submitted, the scale provides four response values (1, 2, 4, and 5). PRobELM \citep{yuan2024probelm} constructs one most-plausible scenario and ten less-plausible alternatives from Wikidata, then evaluates LM rankings of those scenarios. Related tasks assess other units, including candidate clarifications in CLAIRE \citep{anthonio2022claire} and event-based modality categories \citep{pyatkin2021possible}. These resources provide binary labels, pairwise changes, absolute ratings, rankings, or modality categories rather than slot-controlled matched sets.

\paragraph{Controlled and atypical content generation.}
\citet{zhao2024uncommonsense} generate abductive explanations that make unexpected outcomes more likely in context. \citet{li2024longtail} use logical-rule-guided search to generate factually correct but low-confidence inferential statements. \citet{qian2025vgt} use GPT-4o to generate structurally controlled neurolinguistic vignettes with expected and unexpected conditions. More directly, \citet{eichel-schulte-im-walde-2023-dataset} replace two constituents of corpus-derived event triples to produce pseudo-implausible examples, while \citet{tang2023lesslikely} train models to generate relevant but less-likely alternative hypotheses. These methods provide vignette-level structural control, two-constituent perturbation, or one-band generation rather than shared-frame, single-slot control across multiple ordered conditions. To our knowledge, none starts from a root verb, generates all event-slot values, and constructs multiple plausibility targets while holding every non-target slot fixed. This slot-controlled construction setting defines the STRIVE generation contribution rather than graded plausibility generation in general.

\paragraph{Evaluating and refining plausibility.}
\citet{kauf2024logprob} compare sentence log probabilities with direct zero-shot prompting and find log probabilities more reliable for estimating semantic plausibility. \citet{amouyal2024pretesting} test LM ratings as substitutes for human psycholinguistic pretesting and find that even GPT-4 lacks sufficient fine-grained discrimination. VERA \citep{liu2023vera} is a trained general-purpose plausibility estimator for commonsense statements. Using minimally different event pairs, \citet{kauf2023eventknowledge} find that LMs distinguish possible from impossible events more consistently than likely from unlikely events. EWoK \citep{ivanova2025ewok} evaluates conceptual world knowledge more broadly rather than event plausibility alone. Self-Refine \citep{madaan2023selfrefine} uses feedback from the generating model itself, while Cross-Refine \citep{wang-etal-2025-cross} uses feedback from a separate critic model. STRIVE instead uses evaluator verdicts on both condition placement and set-level slot control to iteratively revise a matched four-condition set, while evaluator models and reasoning settings are assessed against human judgments (\S\ref{sec:evaluator}; \S\ref{sec:eval_experiments}).

%% file: appendix/A_rb_gen.tex
\section{RB Generator Prompt}
\label{app:rb_prompt}

The full RB generator prompt is shown in Figure~\ref{fig:v7-agent-part1}. It implements the
five-step procedure described in §\ref{sec:rb} and specifies the
output JSON schema.

\input{figures/rb_prompt/v7-agent_part1}
\input{figures/rb_prompt/v7-agent_part2}
\input{figures/rb_prompt/v7-agent_part3}
\input{figures/rb_prompt/v7-agent_part4}
\input{figures/rb_prompt/v7-agent_part5}
\input{figures/rb_prompt/v7-agent_part6}
\input{figures/rb_prompt/v7-agent_part7}
\input{figures/rb_prompt/v7-agent_part8}
\input{figures/rb_prompt/v7-agent_part9}

\subsection{Design Decisions}
\label{sec:design_decisions}

The five-step procedure embeds several design decisions, each
addressing a specific failure mode observed during prompt iteration.

\paragraph{Scene typicality (Step 1).}
LLMs frequently anchor on a location (for example, a library) and
then force the verb into that location, producing scenes that are
setting-familiar but action-atypical. The result is a gradient
where no agent is genuinely plausible because the underlying scene
is itself implausible. To prevent this, Step 1 includes a
scene-typicality test: substitute \textit{someone} for the agent
and ask whether the resulting sentence describes a routine event.
If not, the scene must be redesigned before agent selection begins.

\paragraph{Core-job verification for PE (Step 2).}
LLMs default to setting-familiar agents (a groundskeeper on a
soccer field) when the action-prototypical agent (a goalie) sits
one inferential step away, producing PE agents who are present in
the location but for whom the action is not part of their job. We
therefore require the PE agent to satisfy a stronger standard than
location-presence: the action must be part of the agent's primary
professional duties. The prompt operationalises this via a removal
test, asking whether removing this agent from the location would
stop $[\textit{verb}][\textit{patient}]$ from happening there.

\paragraph{Multi-dimensional IH overlap (Step 3).}
IH is the hardest of the four conditions to operationalise: the
agent must be a near-miss to PE rather than an obvious mismatch.
We give the model a principled selection rule by requiring overlap
with PE on at least two of five dimensions (occupational domain,
setting, tools, patient type, physical actions). We selected the
two-of-five cutoff by inspecting seed examples during prompt development. It
operationalises a near-miss for this instantiation and can be tuned
for other datasets, target slots, or domains; we do not treat it as
a universal semantic boundary. The four
conditions are first defined by their relation to the scene (PE =
typical doer; PH = present but not typical; IH = related domain
but not present; IE = unrelated); the $\geq 2$ rule is an
additional constraint within the IH cell, so a high-overlap
profession that is action-prototypical for the scene falls in PE,
not IH. For \textit{catch a soccer ball} (PE = goalie, PH =
referee), a hockey player satisfies IH (shares sport domain and
defensive actions, but does not belong on a soccer field); a
fisherman, whose only link to the scene is the verb \textit{catch},
collapses to IE.

\paragraph{Visual identifiability (Step 4).}
STRIVE outputs are intended for downstream image-based assessment,
where the final stimulus is a photograph with no caption or scene
context. An agent whose profession cannot be identified from
appearance alone may satisfy its intended plausibility
condition without providing an experimental signal: the patient
sees a generic person rather than a specific
role. We therefore require each agent to have a profession
recognisable from a photograph against a plain white background,
ruling out professions whose visual presentation collapses to
ordinary civilian clothing (software engineer, freelance writer,
accountant). Pairwise distinctiveness across all six combinations
is verified at the same step.

\paragraph{Pre-output reasoning trace (cross-cutting).}
Per-stimulus rationales capture the model's reasoning about
individual agent choices but do not give the model space to design
the scene holistically, audit overlap dimensions across agents, or
resolve visual conflicts before committing to specific outputs. We
therefore include a free-form scratchpad at the top of the JSON
schema where the entire procedure is worked through before any
final commitments. This trace is what F1 in §\ref{sec:intro}
identifies as the central lever: removing it cuts GPT-5.1's RB GOLD
rate from 28.3\% to 16.7\%, even with per-stimulus rationales
preserved.

%% file: figures/rb_prompt/v7-agent_part1.tex
\begin{figure*}[t]
\centering
\begin{tcolorbox}[width=\textwidth, colback=gray!5, colframe=gray!75,
                  title={RB Generation Prompt (Part 1/9)}]
\begin{Verbatim}[fontsize=\footnotesize]
=== SYSTEM PROMPT ===

You are an expert psycholinguist who understands event cognition, thematic roles, and typicality
effects.
Your task is to generate event sentences that vary systematically in plausibility.
You reason step-by-step before producing output, and you return only valid JSON.

=== USER TEMPLATE ===

TASK
You will be given a verb (root form) at the end of this prompt. Your job is to generate exactly
four event sentences that vary in plausibility for that verb.
Each sentence describes a concrete, observable scene that could be depicted in a single
  photograph.
IMPORTANT: These sentences will later be used to generate images. A human participant will then
view each image and judge whether the depicted event is plausible or not. Therefore, the four
agents must be visually distinguishable from one another in a photograph -- each agent's
  profession
should be identifiable from their appearance alone (clothing, gear, accessories, physique, etc.).

---
EXPERIMENTAL CONTEXT
---
These sentences will be converted into images shown to brain injury patients who must make quick
plausible/implausible judgments. The patients see ONLY the image -- no text, no captions, no
  labels.
Therefore:
  (1) Each agent's profession must be INSTANTLY recognizable from appearance alone.
  (2) The plausibility distinction must be visually obvious and must not depend on subtle
    reasoning.
  (3) The scene must be concrete and unambiguous.
If a profession cannot be identified from a photograph without a caption, do NOT use that
  profession.

---
SENTENCE STRUCTURE
---
Every sentence MUST follow this exact slot order:

  [Agent] [verb] [patient] [instrument] [location]

Slot definitions:
  - Agent: A living human identified by profession or social role (the doer of the action).
  - Verb: A chosen inflected form of the root verb (e.g., "is catching", "catches", "caught").
    Use the SAME inflected form in ALL four sentences.
  - Patient: The object or entity being acted upon.
  - Instrument: The tool, body part, or means used to perform the action (phrased with "with").
  - Location: Where the event takes place (phrased with "in", "on", "at", etc.).

Example: "A goalie is catching a soccer ball with his hands on the soccer field."
  Agent = "A goalie"
  Verb = "is catching"
  Patient = "a soccer ball"
  Instrument = "with his hands"
  Location = "on the soccer field"

---
CORE CONCEPT: THEMATIC FIT
\end{Verbatim}
\end{tcolorbox}
\caption{RB Generation Prompt. Continued in Figure~\ref{fig:v7-agent-part2}.}
\label{fig:v7-agent-part1}
\end{figure*}

%% file: figures/rb_prompt/v7-agent_part2.tex
\begin{figure*}[t]
\centering
\begin{tcolorbox}[width=\textwidth, colback=gray!5, colframe=gray!75,
                  title={RB Generation Prompt (Part 2/9)}]
\begin{Verbatim}[fontsize=\footnotesize]
---
Thematic fit is how naturally a particular AGENT fits as the doer of the described action
(verb + patient + instrument + location). It is about whether this professional, in their
typical work context, would perform this specific action on this patient using this instrument
in this location.

Example: Given the sentence frame "___ is catching a soccer ball with his hands on the soccer
  field":
  - "A goalie" has HIGH thematic fit (this is exactly their job).
  - "A referee" has MODERATE thematic fit (present in the setting, could do it, not their main
    role).
  - "A hockey player" has LOW thematic fit (related sports domain, but wrong sport and wrong
    setting).
  - "A ballerina" has VERY LOW thematic fit (completely unrelated domain).

The four conditions form a gradient of thematic fit from high to low:

  plausible_easy --- plausible_hard --- implausible_hard --- implausible_easy
  (clear yes)        (unsure yes)       (unsure no)          (clear no)
  HIGH fit            MODERATE fit       LOW fit              VERY LOW fit

---
CONDITION DEFINITIONS
---

1) PLAUSIBLE_EASY -- "clear yes" -- HIGH thematic fit
   The agent is the prototypical, most expected professional for performing [verb] on [patient]
   with [instrument] in [location]. The event is physically possible, reasonable, and typical.
   A human would instantly accept this as a normal, everyday event.
   Agent selection: Pick the profession whose job routinely involves exactly this action
   in this setting.

2) PLAUSIBLE_HARD -- "unsure yes" -- MODERATE thematic fit
   The agent is a professional who could plausibly perform [verb] on [patient] with [instrument]
   in [location], but it is NOT their primary role. The event is physically possible and
     reasonable,
   but less typical or less frequent.
   A human would pause but ultimately accept it as possible.
   Agent selection: Pick a profession that is ALREADY PRESENT or COULD NATURALLY BE PRESENT in
   [location], and who has the physical ability to perform [verb] on [patient] with [instrument],
   but for whom this action is secondary or occasional rather than central to their role.
   KEY TEST: Can you easily imagine a specific, realistic scenario where this professional does
   this? If yes, it qualifies.

3) IMPLAUSIBLE_HARD -- "unsure no" -- LOW thematic fit
   The agent is a professional from a RELATED but WRONG domain. There IS semantic overlap between
   the agent's profession and the plausible_easy agent's profession, but the agent still
   does NOT belong in [location] performing [verb] on [patient] with [instrument].
   The event is unreasonable and atypical, but the domain similarity makes a human hesitate
   before rejecting it.

   CRITICAL -- MULTI-DIMENSIONAL OVERLAP REQUIREMENT:
   The implausible_hard agent must share overlap with the plausible_easy agent on AT LEAST TWO
   of the following five dimensions:
     (a) Occupational domain (e.g., both belong to the same broad professional field)
     (b) Typical work setting/environment (e.g., both typically work in similar environments)
     (c) Tools or instruments routinely used (e.g., both routinely use similar types of tools)
     (d) Type of patient/object acted upon (e.g., both act on similar types of objects or
       entities)
     (e) Physical actions routinely performed (e.g., both routinely perform similar physical
       actions)
\end{Verbatim}
\end{tcolorbox}
\caption{RB Generation Prompt (continued from previous page).}
\label{fig:v7-agent-part2}
\end{figure*}

%% file: figures/rb_prompt/v7-agent_part3.tex
\begin{figure*}[t]
\centering
\begin{tcolorbox}[width=\textwidth, colback=gray!5, colframe=gray!75,
                  title={RB Generation Prompt (Part 3/9)}]
\begin{Verbatim}[fontsize=\footnotesize]

   A single dimension of overlap (e.g., both use sharp tools) is NOT SUFFICIENT. If the only
   connection is through the verb itself, the agent belongs in implausible_easy, not here.
   NEGATIVE TEST: Remove the shared verb from consideration. Is there STILL a reason to associate
   this agent with the scene? If not, the overlap is too thin.

   Agent selection: Pick a profession that shares at least two of the above dimensions with the
   plausible_easy agent but who would NOT perform THIS specific action in THIS specific setting.
   KEY TEST: Does this agent make you briefly think "wait, maybe..." before you conclude "no"?
   If yes, it qualifies.

4) IMPLAUSIBLE_EASY -- "clear no" -- VERY LOW thematic fit
   The agent is a professional from a COMPLETELY UNRELATED domain. There is no semantic connection
   between the agent's profession and the action, patient, instrument, or location.
   The event is unreasonable and deeply atypical.
   A human would instantly reject it.
   Agent selection: Pick a profession that has NO topical, domain, or contextual overlap with any
   part of the event. The agent's professional world should be maximally distant from the scene.

---
HARD CONSTRAINTS
---

AGENT RULES:
- Every agent MUST be a living human identified by a real profession or social role.
  Valid examples: any real-world profession or social role (e.g., "a [profession]").
- Age-based social roles with clear visual markers are also acceptable when they carry
  distinct role expectations. Valid examples: "a schoolchild" (school uniform, backpack,
  small stature), "a retiree" or "an elderly person" (gray hair, glasses, possibly a cane).
  These roles are useful because people have schematic expectations about what a schoolchild
  or a retiree would or would not do, which creates natural thematic fit variation.
- IMPORTANT: Plausibility must always stem from ROLE-ACTION MISMATCH (whether this role
  typically performs this action), NOT from physical incapability. Do not use agents whose
  implausibility comes from being physically unable to perform the action (e.g., an infant
  lifting heavy equipment). The question is "does this role fit this event?" not "can this
  person physically do this?"
- FORBIDDEN agents: animals, objects, fictional beings, statues, robots, corpses, body parts,
  descriptions of states (e.g., "a sleeping person"), physical descriptors without a role
  (e.g., "a tall person", "an overweight person"), or any non-human entity.
- Assume each agent wears clothing and gear appropriate to their profession or social role.

PROFESSION IDENTIFIABILITY RULES:
- Every agent MUST have a profession that is visually identifiable from a photograph WITHOUT
  any caption, label, or scene context. A naive viewer seeing ONLY the person in their typical
  professional attire must be able to correctly guess their profession.
- A profession is visually identifiable if it has at least one of the following:
  (a) A dedicated uniform or standardized work attire distinct from everyday civilian clothing.
  (b) Profession-specific gear or equipment that is worn or carried as part of the role.
  (c) A highly distinctive physical presentation strongly associated with the role.
- A profession is NOT visually identifiable if the person would look like a generic civilian
  in everyday clothing. The test: Imagine this person standing alone against a plain white
  background -- no scene context, no caption, no other people. Could a viewer correctly guess
  their profession from appearance alone? If not, do NOT use that profession.
- When in doubt, prefer professions with uniforms or standardized professional attire over
  those with informal or variable dress codes.
\end{Verbatim}
\end{tcolorbox}
\caption{RB Generation Prompt (continued from previous page).}
\label{fig:v7-agent-part3}
\end{figure*}

%% file: figures/rb_prompt/v7-agent_part4.tex
\begin{figure*}[t]
\centering
\begin{tcolorbox}[width=\textwidth, colback=gray!5, colframe=gray!75,
                  title={RB Generation Prompt (Part 4/9)}]
\begin{Verbatim}[fontsize=\footnotesize]

VISUAL DISTINCTIVENESS RULES:
- All four agents must be visually distinguishable from one another in a photograph.
- Each agent's profession should be identifiable by their typical professional appearance:
  clothing, uniform, gear, accessories, protective equipment, or other visual markers.
- STRONGLY PREFER professions that have a recognizable "look" -- those with distinctive
  uniforms, professional gear, or role-specific attire that a viewer could identify without
  any caption or context. The more visually iconic the profession, the better.
  Avoid: professions that look like generic civilians in everyday clothing, such as
        "a freelance writer", "a software engineer", "a real estate agent", "an accountant".
- Pairs of agents that would look nearly identical in a photo are NOT acceptable.
  If two agents would look nearly identical in a photograph due to similar workwear, replace
  one with a profession that has a more distinctive visual identity while preserving thematic fit.
- When two professions share the same broad domain, ensure they differ in at least one
  strong visual cue (uniform type, headgear, tools they carry, body build, etc.).

CONSISTENCY RULES:
- The verb form, patient, instrument, and location MUST be IDENTICAL in all four sentences.
- ONLY the agent changes between conditions.

IMAGEABILITY:
- Each sentence must depict a scene that a photographer could capture in one still image.
- Avoid abstract, metaphorical, or unobservable events.

---
PROCEDURE (follow these six steps in order)
---

STEP 1 -- DESIGN THE SCENE
Choose a naturalistic, concrete scene for the root verb.
Select a specific verb form, patient, instrument, and location that together create a vivid,
everyday scenario. The scene should be one where a clear prototypical agent exists.

STEP 1b -- VERIFY SCENE TYPICALITY
Before selecting any agents, verify that the scene itself supports a prototypical event.
Construct a test sentence by replacing the agent with "someone":
  "Someone [verb_form] [patient] [instrument] [location]."
Ask: Does this sentence describe a ROUTINE, EVERYDAY event that a person would recognize as
typical and unsurprising? Would a person hearing this sentence think "yes, that happens
  regularly"?

If the answer is NO -- the scene is atypical regardless of the agent (e.g., "Someone is catching
a soccer ball with his hands in a library" -- catching soccer balls in a library is not routine
for anyone) -- then REDESIGN the scene. Choose a different patient, instrument, or location where
the verb represents a routine, everyday activity with a clear prototypical agent.

This step prevents a common failure mode: choosing a location first (e.g., library) and then
forcing a setting-familiar agent (e.g., librarian) into an action that is not actually typical
in that setting. The scene must be action-typical, not just setting-familiar.

Only proceed to Step 2 after the "someone" test passes.
\end{Verbatim}
\end{tcolorbox}
\caption{RB Generation Prompt (continued from previous page).}
\label{fig:v7-agent-part4}
\end{figure*}

%% file: figures/rb_prompt/v7-agent_part5.tex
\begin{figure*}[t]
\centering
\begin{tcolorbox}[width=\textwidth, colback=gray!5, colframe=gray!75,
                  title={RB Generation Prompt (Part 5/9)}]
\begin{Verbatim}[fontsize=\footnotesize]

STEP 2 -- SELECT FOUR AGENTS ALONG THE THEMATIC-FIT GRADIENT
Working from the scene you designed:
  a) plausible_easy agent: Who is THE most expected professional performing [verb] on [patient]
     with [instrument] in [location]? (prototypical, high thematic fit)
     CORE-JOB VERIFICATION: Confirm that [verb] [patient] is part of this agent's PRIMARY
     professional duties -- not merely something they could do, but something they are PAID or
     EXPECTED to do routinely. Ask: "If this agent were removed from [location], would [verb]
     [patient] stop happening there?" If the answer is no, this agent is setting-prototypical
     (familiar with the location) but not action-prototypical (the action is not their core job).
     A setting-prototypical agent belongs in plausible_hard, not plausible_easy.
     Example: A groundskeeper on a soccer field is setting-prototypical, but "catching
     a soccer ball" is not a groundskeeper's job. The groundskeeper belongs in plausible_hard
       at best.
  b) plausible_hard agent: Who else might do this in [location], even if it is not their main
     job? (atypical but naturally present, moderate thematic fit)
  c) implausible_hard agent: Who SEEMS related due to domain overlap but does NOT belong in this
     specific scene? This agent must share overlap on AT LEAST TWO of the five dimensions
     (occupational domain, setting, tools, patient type, physical actions) with the plausible_easy
     agent. Apply the negative test: removing the shared verb, is there still a reason to
     associate this agent with the scene?
  d) implausible_easy agent: Who has ZERO connection to any part of this scene?
     (maximally distant profession, very low thematic fit)

STEP 3 -- CHECK VISUAL DISTINCTIVENESS AND IDENTIFIABILITY
First, for each agent individually, ask: Is this profession visually identifiable from a
photograph against a plain white background -- no scene context, no caption, no other people?
Does this agent have a uniform, professional gear, or distinctive attire that makes their
profession obvious? If not, replace the agent with a profession that has stronger visual
identity while preserving its thematic-fit level.

Then, for each pair of agents, ask: Would these two look different in a photograph?
Consider their typical professional attire, gear, accessories, and physical presentation.
  - Can a viewer tell plausible_easy from plausible_hard by appearance? If not, replace one.
  - Can a viewer tell implausible_hard from implausible_easy by appearance? If not, replace one.
  - Can a viewer tell ANY two agents apart? If any pair looks identical, replace one agent
    with a profession that has a more distinctive visual identity while preserving its
    thematic-fit level.
If you make replacements, re-verify that thematic fit is still correct for the new agent.

STEP 4 -- COMPOSE FOUR SENTENCES
Insert each agent into the shared sentence frame:
  [Agent] [verb form] [patient] [instrument] [location].

STEP 5 -- VALIDATE
Check every item below before producing output:
  - All four agents are living humans with identifiable professions? YES/NO
  - All four agents are visually identifiable from a photograph without captions? YES/NO
  - Verb form, patient, instrument, and location identical across all four sentences? YES/NO
  - plausible_easy: Would a person INSTANTLY say "yes, that's normal"? YES/NO
  - Scene typicality: Does "Someone [verb] [patient] [instrument] [location]" sound routine?
    YES/NO
  - plausible_easy CORE-JOB: Is [verb] [patient] part of this agent's primary job description?
    YES/NO
  - plausible_easy: Is the agent action-prototypical (not just setting-familiar)? YES/NO
  - plausible_hard: Would a person PAUSE then say "yes, I suppose that could happen"? YES/NO
  - implausible_hard: Would a person HESITATE then say "no, that doesn't fit"? YES/NO
  - implausible_hard: Does this agent share at least TWO overlap dimensions with plausible_easy?
    YES/NO
  - implausible_hard: Removing the shared verb, is there still a reason to associate this agent
    with the scene? YES/NO
  - implausible_easy: Would a person INSTANTLY say "no, that's absurd"? YES/NO
\end{Verbatim}
\end{tcolorbox}
\caption{RB Generation Prompt (continued from previous page).}
\label{fig:v7-agent-part5}
\end{figure*}

%% file: figures/rb_prompt/v7-agent_part6.tex
\begin{figure*}[t]
\centering
\begin{tcolorbox}[width=\textwidth, colback=gray!5, colframe=gray!75,
                  title={RB Generation Prompt (Part 6/9)}]
\begin{Verbatim}[fontsize=\footnotesize]
  - Ordering is correct: plausible_easy > plausible_hard >> implausible_hard > implausible_easy?
    YES/NO
  - Each sentence is imageable as a single photograph? YES/NO
  - All four agents are visually distinguishable from each other in a photograph? YES/NO
  - Each agent's profession can be identified from their typical appearance? YES/NO
If any check fails, revise before outputting.

STEP 6 -- OUTPUT
Return the JSON object specified in OUTPUT FORMAT below.

---
WORKED EXAMPLE (root verb = "catch")
---

Scene design:
  verb_form = "is catching", patient = "a soccer ball",
  instrument = "with his hands", location = "on the soccer field"

Scene typicality check:
  "Someone is catching a soccer ball with his hands on the soccer field."
  Is this routine? YES -- catching soccer balls on a field is a typical, everyday sporting event.
  Proceed to agent selection.

Agent selection along thematic-fit gradient:
  a) plausible_easy -> "A goalie":
     Catching the ball is the goalie's primary job on the soccer field. Prototypical agent.
       HIGH fit.
     Visual: goalie jersey, gloves, distinct from other players.
  b) plausible_hard -> "A referee":
     Referees are present on the soccer field and might catch a ball occasionally (e.g., ball
     thrown to them, preventing a deflection). Not their main role, but realistic. MODERATE fit.
     Visual: black referee uniform, whistle -- clearly different from a goalie.
  c) implausible_hard -> "A hockey player":
     Hockey is a related team sport (athletes, goals, ball/puck), creating domain overlap that
     causes momentary confusion. But a hockey player does not belong on a soccer field. LOW fit.
     Overlap dimensions: (a) occupational domain -- both are team sport athletes; (e) physical
     actions -- both catch/block projectiles aimed at a goal. That is 2 dimensions = sufficient.
     Negative test: Removing "catching" from consideration, is there still a reason to associate
     a hockey player with a soccer field? Yes -- both are team sports with goals, balls/pucks,
     fields/rinks, and athletic roles. The association persists beyond the verb.
     Visual: hockey jersey, pads, helmet -- distinct from both goalie and referee.
  d) implausible_easy -> "A ballerina":
     Ballet has zero connection to soccer, team sports, or outdoor fields. No semantic overlap
     with any part of the scene. VERY LOW fit.
     Visual: tutu, pointe shoes, hair in bun -- maximally distinct from all others.

Visual identifiability check:
  - Goalie: goalie jersey + gloves = IDENTIFIABLE (distinct sports uniform)
  - Referee: black uniform + whistle = IDENTIFIABLE (recognizable official attire)
  - Hockey player: hockey gear + pads + helmet = IDENTIFIABLE (sport-specific equipment)
  - Ballerina: tutu + pointe shoes + bun = IDENTIFIABLE (iconic performance attire)

Visual distinctiveness check (pairwise):
  - Goalie (goalie jersey + gloves) vs Referee (black uniform + whistle): DISTINCT
  - Hockey player (hockey gear + pads) vs Ballerina (tutu + pointe shoes): DISTINCT
  - All six pairwise comparisons: DISTINCT
\end{Verbatim}
\end{tcolorbox}
\caption{RB Generation Prompt (continued from previous page).}
\label{fig:v7-agent-part6}
\end{figure*}

%% file: figures/rb_prompt/v7-agent_part7.tex
\begin{figure*}[t]
\centering
\begin{tcolorbox}[width=\textwidth, colback=gray!5, colframe=gray!75,
                  title={RB Generation Prompt (Part 7/9)}]
\begin{Verbatim}[fontsize=\footnotesize]

Output:
{
  "reasoning": "STEP 1 - SCENE DESIGN: For 'catch', a natural everyday scene is a soccer field
    where catching is a routine action. verb_form='is catching', patient='a soccer ball',
    instrument='with his hands', location='on the soccer field'. The goalie is the clear
    prototypical doer.\n\nSTEP 1b - SCENE TYPICALITY CHECK: 'Someone is catching a soccer ball
    with his hands on the soccer field.' Is this a routine, unsurprising event? YES -- this
    happens every match. The scene is action-typical, not just setting-familiar. Proceed to
    agent selection.\n\nSTEP 2 - AGENT SELECTION:\n(a) plausible_easy = A goalie. CORE-JOB
    VERIFICATION: catching shots on goal is the goalie's primary professional duty -- they are
    paid to do exactly this. REMOVAL TEST: if the goalie is removed from the field, catching
    shots at the goal stops happening. Goalie is action-prototypical, not just setting-familiar.
    HIGH fit.\n(b) plausible_hard = A referee. Already present on the field and could
    realistically catch a ball thrown to them by a player or block a stray deflection. Not their
    main role but easy to imagine. MODERATE fit.\n(c) implausible_hard = A hockey player.
    Overlap dimensions with goalie: (1) occupational_domain -- both are team sport athletes; (2)
    physical_actions -- both catch or block projectiles aimed at a goal. Two dimensions =
    sufficient. Negative test: remove the verb 'catching' from consideration. Is there still a
    reason to associate a hockey player with a soccer field? Yes -- both are team sports played
    on a field/rink with goals, athletic roles, and competitive structure. The association
    persists beyond the verb. LOW fit.\n(d) implausible_easy = A ballerina. Ballet has zero
    connection to soccer, team sports, or outdoor fields. No semantic overlap on any dimension.
    VERY LOW fit.\n\nSTEP 3 - VISUAL CHECK:\nIdentifiability (each agent against a plain white
    background):\n- Goalie: goalie jersey + padded gloves = IDENTIFIABLE.\n- Referee: black
    striped uniform + whistle = IDENTIFIABLE.\n- Hockey player: hockey jersey + shoulder/shin
    pads + helmet = IDENTIFIABLE.\n- Ballerina: tutu + pointe shoes + bun =
    IDENTIFIABLE.\nPairwise (6 pairs): all distinct. Goalie vs referee differ in jersey color
    and gloves; hockey player vs ballerina are maximally different; cross-pairs are obviously
    distinct. No replacements needed.\n\nSTEP 5 - VALIDATION: All checks pass -- four living
    human professions, all visually identifiable, slot consistency across all four sentences,
    scene typicality confirmed, plausible_easy passes core-job and action-prototypicality,
    implausible_hard has 2 overlap dimensions and survives the negative test, gradient ordering
    is correct, all four sentences imageable as single photographs, and pairwise visual
    distinctiveness holds for all six pairs.",
  "scene": {
    "verb_form": "is catching",
    "patient": "a soccer ball",
    "instrument": "with his hands",
    "location": "on the soccer field"
  },
  "agents": {
    "plausible_easy": {
      "agent": "A goalie",
      "thematic_fit": "HIGH",
      "rationale": "Catching the ball is the goalie's primary job on the soccer field.
        Prototypical agent for this event.",
      "visual_markers": "Goalie jersey in a bright color, padded goalie gloves, athletic
        shorts, cleats."
    },
    "plausible_hard": {
      "agent": "A referee",
      "thematic_fit": "MODERATE",
      "rationale": "Referees are present on the field and could catch the ball in unusual
        but realistic situations. Not their main role.",
      "visual_markers": "Black referee uniform with vertical stripes, whistle around neck,
        no gloves."
    },
    "implausible_easy": {
      "agent": "A ballerina",
      "thematic_fit": "VERY LOW",
      "rationale": "Ballet is an unrelated domain with zero semantic overlap to soccer or
\end{Verbatim}
\end{tcolorbox}
\caption{RB Generation Prompt (continued from previous page).}
\label{fig:v7-agent-part7}
\end{figure*}

%% file: figures/rb_prompt/v7-agent_part8.tex
\begin{figure*}[t]
\centering
\begin{tcolorbox}[width=\textwidth, colback=gray!5, colframe=gray!75,
                  title={RB Generation Prompt (Part 8/9)}]
\begin{Verbatim}[fontsize=\footnotesize]
        team sports.",
      "visual_markers": "White or pink tutu, pointe shoes, hair in a tight bun, slender build."
    },
    "implausible_hard": {
      "agent": "A hockey player",
      "thematic_fit": "LOW",
      "rationale": "Hockey is a related team sport (athletes, goals, ball/puck), creating
        domain overlap, but a hockey player does not belong on a soccer field.",
      "overlap_dimensions": ["occupational_domain", "physical_actions"],
      "visual_markers": "Hockey jersey, bulky shoulder and shin pads, hockey helmet with
        visor, ice skates."
    }
  },
  "sentences": {
    "plausible_easy": "A goalie is catching a soccer ball with his hands on the soccer field.",
    "plausible_hard": "A referee is catching a soccer ball with his hands on the soccer field.",
    "implausible_easy": "A ballerina is catching a soccer ball with his hands on the soccer
      field.",
    "implausible_hard": "A hockey player is catching a soccer ball with his hands on the
      soccer field."
  }
}
---
OUTPUT FORMAT (STRICT)
---

Return ONLY the following JSON object. No markdown fences. No text outside the JSON.

{
  "reasoning": "<Use this field as a scratchpad. Work through the entire 6-step PROCEDURE here:
    design the scene, verify scene typicality with the 'someone' test, select all four agents
    with thematic fit reasoning (including the core-job verification and removal test for
    plausible_easy, and the multi-dimensional overlap audit plus negative test for
    implausible_hard), check visual identifiability and pairwise distinctiveness, compose
    sentences, and run all validation checks. Write freely -- this is your workspace to think
    through the problem before committing to the output below.>",
  "scene": {
    "verb_form": "<chosen inflected form of the root verb>",
    "patient": "<patient phrase>",
    "instrument": "<instrument phrase starting with 'with'>",
    "location": "<location phrase starting with a preposition>"
  },
  "agents": {
    "plausible_easy": {
      "agent": "<agent phrase>",
      "thematic_fit": "HIGH",
      "rationale": "<1-2 sentences: why this agent is prototypical for this event>",
      "visual_markers": "<brief description of this agent's distinctive professional appearance>"
    },
    "plausible_hard": {
      "agent": "<agent phrase>",
      "thematic_fit": "MODERATE",
      "rationale": "<1-2 sentences: why this agent is plausible but atypical>",
      "visual_markers": "<brief description of this agent's distinctive professional appearance>"
    },
    "implausible_easy": {
      "agent": "<agent phrase>",
      "thematic_fit": "VERY LOW",
      "rationale": "<1-2 sentences: why this agent is from a completely unrelated domain>",
      "visual_markers": "<brief description of this agent's distinctive professional appearance>"
    },
    "implausible_hard": {
\end{Verbatim}
\end{tcolorbox}
\caption{RB Generation Prompt (continued from previous page).}
\label{fig:v7-agent-part8}
\end{figure*}

%% file: figures/rb_prompt/v7-agent_part9.tex
\begin{figure*}[t]
\centering
\begin{tcolorbox}[width=\textwidth, colback=gray!5, colframe=gray!75,
                  title={RB Generation Prompt (Part 9/9)}]
\begin{Verbatim}[fontsize=\footnotesize]
      "agent": "<agent phrase>",
      "thematic_fit": "LOW",
      "rationale": "<1-2 sentences: why this agent is from a related but wrong domain>",
      "overlap_dimensions": ["<list the 2+ overlapping dimensions from: occupational_domain,
        typical_setting, tools_instruments, patient_object_type, physical_actions>"],
      "visual_markers": "<brief description of this agent's distinctive professional appearance>"
    }
  },
  "sentences": {
    "plausible_easy": "<full sentence>",
    "plausible_hard": "<full sentence>",
    "implausible_easy": "<full sentence>",
    "implausible_hard": "<full sentence>"
  }
}

---
VERB INPUT (process this verb now)
---
The root verb for this generation is: "{verb}"

Follow the PROCEDURE above for this verb. Return ONLY the JSON object. No other text.
\end{Verbatim}
\end{tcolorbox}
\caption{RB Generation Prompt (continued from previous page).}
\label{fig:v7-agent-part9}
\end{figure*}

%% file: appendix/B_r2v_gen.tex
\section{R2V Generator Prompt}
\label{app:r2v_prompt}

Since the prompts are very long and occupy many pages, we show the portion of the prompt which dictates the verbalized sampling~\citep{zhang2026verbalized} in Figure~\ref{fig:v7-agent-verbalized-sampling-part5}. The complete prompt would be released upon publication along with the code. 

\input{figures/r2v_prompt/v7-agent-verbalized-sampling_part5}

\section{Iterative Refine Prompt}
\label{app:ir_prompt}

The common refinement prompt contains the same task definitions and validation rubric as RB, with two additions: an instruction to rectify rather than regenerate (Figure~\ref{fig:v7-agent-iterative-part1}) and a feedback-history placeholder (Figure~\ref{fig:v7-agent-iterative-part13}). At each iteration, this field contains accumulated compact, instance-specific diagnoses rather than the evaluator rubric or full output. The previously generated output is supplied separately. An abridged diagnosis has the following structure:

{\color{blue}
\begin{Verbatim}[fontsize=\scriptsize]
AGENTS:
  plausible_hard:
    agent: "A librarian"
    verdict: INCORRECT
    actual_condition: implausible_easy
    issues: Librarian has no natural presence in a home
    living room.
  implausible_hard:
    agent: "A carpenter"
    verdict: CORRECT
    overlap_count: 2
    overlap_dimensions:
      [typical_setting, patient_object_type]
GRADIENT:
  correctly_ordered: NO
SCENE:
  verdict: NEEDS_REDESIGN
\end{Verbatim}
}
The overlap fields report findings for a specific generated agent; they do not restate the IH definition.

The complete prompt would be released upon publication along with the code.

\input{figures/v7-agent-iterative_part1}
\input{figures/v7-agent-iterative_part13}

%% file: figures/r2v_prompt/v7-agent-verbalized-sampling_part5.tex
\begin{figure*}[t]
\centering
\begin{tcolorbox}[width=\textwidth, colback=gray!5, colframe=gray!75,
                  title={R2V Generation Prompt (Partial)}]
\begin{Verbatim}[fontsize=\footnotesize]

Distribution guidance -- aim for approximately:
  - 3-4 candidates with probability 0.70-1.00 (prototypical agents, HIGH fit)
  - 5-6 candidates with probability 0.35-0.69 (plausible but atypical, MODERATE fit)
  - 5-6 candidates with probability 0.10-0.34 (related domain but wrong, LOW fit)
  - 3-4 candidates with probability 0.00-0.09 (completely unrelated, VERY LOW fit)

IMPORTANT DIVERSITY RULES FOR CANDIDATE GENERATION:
  - Do NOT cluster candidates within one professional field. Spread across diverse domains:
    medical, sports, trades, arts, military, food service, education, emergency services,
    religious, legal, scientific, transportation, agriculture, etc.
  - Every candidate MUST be visually identifiable from a photograph against a plain white
    background -- no scene context, no caption, no other people. Do NOT include any profession
    that looks like a generic civilian.
  - Think carefully about the probability estimates. A probability of 0.50 means a human
    would be genuinely uncertain whether this agent fits the event. A probability of 0.25
    means a human would lean toward "no" but see why someone might think otherwise.

STEP 3 -- SELECT THE OPTIMAL FOUR-AGENT COMBINATION
From your 20 candidates, select exactly four agents -- one for each plausibility condition.
Optimize for GRADIENT SPACING: the four selected agents should be well-separated in their
thematic fit probabilities, not clustered together.

Selection criteria:
  a) plausible_easy: Select the candidate with the highest thematic fit probability.
     Target range: 0.80-1.00. This must be THE prototypical agent for this event.
     CORE-JOB VERIFICATION: Confirm that [verb] [patient] is part of this agent's PRIMARY
     professional duties -- not merely something they could do, but something they are PAID or
     EXPECTED to do routinely. Ask: "If this agent were removed from [location], would [verb]
     [patient] stop happening there?" If the answer is no, this agent is setting-prototypical
     (familiar with the location) but not action-prototypical (the action is not their core job).
     A setting-prototypical agent belongs in plausible_hard, not plausible_easy.
     Example: A groundskeeper on a soccer field is setting-prototypical, but "catching
     a soccer ball" is not a groundskeeper's job. The groundskeeper belongs in plausible_hard
       at best.
  b) plausible_hard: Select a candidate with probability in the 0.40-0.65 range.
     This agent must be naturally present or could naturally be present in the location,
     and the action must be secondary or occasional for them, not their primary role.
  c) implausible_hard: Select a candidate with probability in the 0.15-0.35 range.
     This agent MUST share overlap with the plausible_easy agent on AT LEAST TWO of the
     five overlap dimensions (occupational domain, typical setting, tools/instruments,
     patient/object type, physical actions). Apply the NEGATIVE TEST: removing the shared
     verb, is there STILL a reason to associate this agent with the scene?
  d) implausible_easy: Select the candidate with the lowest thematic fit probability.
     Target range: 0.00-0.05. This must be from a COMPLETELY UNRELATED domain.

STEP 4 -- CHECK COMBINATION QUALITY
Verify the selected four agents satisfy ALL of the following:
  - All four are living humans with identifiable professions? YES/NO
  - All four are visually identifiable from a photograph against a plain white background? YES/NO
  - All four are visually distinguishable from each other in a photograph? YES/NO
  - Scene typicality: Does "Someone [verb] [patient] [instrument] [location]" sound routine?
    YES/NO
  - plausible_easy: Would a person INSTANTLY say "yes, that's normal"? YES/NO
  - plausible_easy CORE-JOB: Is [verb] [patient] part of this agent's primary job description?
    YES/NO
  - plausible_easy: Is the agent action-prototypical (not just setting-familiar)? YES/NO
  - plausible_hard: Would a person PAUSE then say "yes, I suppose that could happen"? YES/NO
  - implausible_hard: Would a person HESITATE then say "no, that doesn't fit"? YES/NO
  - implausible_hard: Does this agent share at least TWO overlap dimensions with plausible_easy?
    YES/NO
  - implausible_hard: Removing the shared verb, is there still a reason to associate this agent
\end{Verbatim}
\end{tcolorbox}
\caption{R2V Generation Prompt (Partial).}
\label{fig:v7-agent-verbalized-sampling-part5}
\end{figure*}

%% file: figures/v7-agent-iterative_part1.tex
\begin{figure*}[t]
\centering
\begin{tcolorbox}[width=\textwidth, colback=gray!5, colframe=gray!75,
                  title={Iterative Prompt (Instruction)}]
\begin{Verbatim}[fontsize=\footnotesize]
=== SYSTEM PROMPT ===

You are an expert psycholinguist who understands event cognition, thematic roles, and typicality
effects.
Your task is to revise a previously generated set of event sentences based on evaluator feedback.
You carefully analyze each criticism and make targeted fixes while preserving what already works.
You reason step-by-step before producing output, and you return only valid JSON.

=== USER TEMPLATE ===

TASK
You will be given a verb (root form), a previous generation attempt, and evaluator feedback
at the end of this prompt. Your job is to revise the previous output to fix ALL identified
issues while preserving anything that was correct.
Each sentence describes a concrete, observable scene that could be depicted in a single
  photograph.
IMPORTANT: These sentences will later be used to generate images. A human participant will then
view each image and judge whether the depicted event is plausible or not. Therefore, the four
agents must be visually distinguishable from one another in a photograph -- each agent's
  profession
should be identifiable from their appearance alone (clothing, gear, accessories, physique, etc.).

---
EXPERIMENTAL CONTEXT
---
These sentences will be converted into images shown to brain injury patients who must make quick
plausible/implausible judgments. The patients see ONLY the image -- no text, no captions, no
  labels.
Therefore:
  (1) Each agent's profession must be INSTANTLY recognizable from appearance alone.
  (2) The plausibility distinction must be visually obvious and must not depend on subtle
    reasoning.
  (3) The scene must be concrete and unambiguous.
If a profession cannot be identified from a photograph without a caption, do NOT use that
  profession.

---
SENTENCE STRUCTURE
---
Every sentence MUST follow this exact slot order:

  [Agent] [verb] [patient] [instrument] [location]

Slot definitions:
  - Agent: A living human identified by profession or social role (the doer of the action).
  - Verb: A chosen inflected form of the root verb (e.g., "is catching", "catches", "caught").
    Use the SAME inflected form in ALL four sentences.
  - Patient: The object or entity being acted upon.
  - Instrument: The tool, body part, or means used to perform the action (phrased with "with").
  - Location: Where the event takes place (phrased with "in", "on", "at", etc.).

Example: "A goalie is catching a soccer ball with his hands on the soccer field."
  Agent = "A goalie"
  Verb = "is catching"
  Patient = "a soccer ball"
  Instrument = "with his hands"
  Location = "on the soccer field"

---
CORE CONCEPT: THEMATIC FIT
\end{Verbatim}
\end{tcolorbox}
\caption{Iterative Prompt (Instruction).}
\label{fig:v7-agent-iterative-part1}
\end{figure*}

%% file: figures/v7-agent-iterative_part13.tex
\begin{figure*}[t]
\centering
\begin{tcolorbox}[width=\textwidth, colback=gray!5, colframe=gray!75,
                  title={Iterative Prompt (Feedback placeholder)}]
\begin{Verbatim}[fontsize=\footnotesize]
      "agent": "<agent phrase>",
      "thematic_fit": "VERY LOW",
      "rationale": "<1-2 sentences: why this agent is from a completely unrelated domain>",
      "visual_markers": "<brief description of this agent's distinctive professional appearance>",
      "changed": false
    },
    "implausible_hard": {
      "agent": "<agent phrase>",
      "thematic_fit": "LOW",
      "rationale": "<1-2 sentences: why this agent is from a related but wrong domain>",
      "overlap_dimensions": ["<list the 2+ overlapping dimensions from: occupational_domain,
        typical_setting, tools_instruments, patient_object_type, physical_actions>"],
      "visual_markers": "<brief description of this agent's distinctive professional appearance>",
      "changed": true
    }
  },
  "sentences": {
    "plausible_easy": "<full sentence>",
    "plausible_hard": "<full sentence>",
    "implausible_easy": "<full sentence>",
    "implausible_hard": "<full sentence>"
  }
}

Note on the 'changed' field: it is true for any agent that differs from the corresponding
agent in the previous output, and false for any agent preserved unchanged. The true/false
values shown above are illustrative only -- set them to reflect your actual changes.

---
REVISION INPUT (process this now)
---

Root verb: "{verb}"

PREVIOUS OUTPUT:
{previous_output}

EVALUATOR FEEDBACK:
{evaluator_feedback}

Follow the REVISION PROCEDURE (Steps R1-R4) then the GENERATION PROCEDURE (Steps 1-6).
Return ONLY the JSON object. No other text.
\end{Verbatim}
\end{tcolorbox}
\caption{Iterative Prompt (Feedback placeholder).}
\label{fig:v7-agent-iterative-part13}
\end{figure*}

%% file: appendix/C_evaluator.tex
\section{Evaluator Prompt}
\label{app:evaluator_prompt}
The STRIVE evaluator assesses each generated set across five stages in fixed order:

\begin{enumerate}
  \item \textbf{Agent constraint validity.} Checks that all four
    agents are living humans with identifiable professions, that
    the verb form, patient, instrument, and location are identical
    across sentences, and that implausibility stems from role-action
    misfit rather than physical incapability.
  \item \textbf{Per-condition plausibility correctness.} For each
    condition, the evaluator applies the adversarial counter-argument
    check (§\ref{sec:evaluator}) and, for IH, the overlap audit.
    Verdicts are \textsc{correct}, \textsc{borderline}, or
    \textsc{incorrect}; \textsc{incorrect} verdicts specify which
    condition the agent actually belongs in.
  \item \textbf{Gradient ordering.} Checks that the overall
    plausibility ordering PE $>$ PH $\gg$ IH $>$ IE is maintained
    and that no adjacent conditions are collapsed.
  \item \textbf{Visual evaluation.} First rates each agent's
    profession identifiability (\textsc{identifiable} /
    \textsc{ambiguous} / \textsc{unidentifiable}); then checks all
    six pairwise agent combinations for visual distinguishability.
    Verdict: \textsc{all\_distinct}, \textsc{mostly\_distinct}, or
    \textsc{problematic}.
  \item \textbf{Scene assessment.} Checks instrument correctness,
    prototype clarity, imageability, and gradient supportability.
    For each \textsc{incorrect} or \textsc{borderline} condition,
    the evaluator states whether a viable replacement agent exists
    for the current scene; if not, the scene itself is the root
    cause of failure.
    Verdict: \textsc{good}, \textsc{agent\_fixable}, or
    \textsc{needs\_redesign}.
\end{enumerate}

The full evaluator prompt is shown in Figure~\ref{fig:evaluator-v3.1-agent-part1}.

\input{figures/evaluator/evaluator-v3.1-agent_part1}
\input{figures/evaluator/evaluator-v3.1-agent_part2}
\input{figures/evaluator/evaluator-v3.1-agent_part3}
\input{figures/evaluator/evaluator-v3.1-agent_part4}
\input{figures/evaluator/evaluator-v3.1-agent_part5}
\input{figures/evaluator/evaluator-v3.1-agent_part6}
\input{figures/evaluator/evaluator-v3.1-agent_part7}
\input{figures/evaluator/evaluator-v3.1-agent_part8}

%% file: figures/evaluator/evaluator-v3.1-agent_part1.tex
\begin{figure*}[t]
\centering
\begin{tcolorbox}[width=\textwidth, colback=gray!5, colframe=gray!75,
                  title={Evaluator Prompt (Part 1/8)}]
\begin{Verbatim}[fontsize=\footnotesize]
=== SYSTEM PROMPT ===

You are an expert psycholinguist and cognitive scientist who evaluates event plausibility
sentences.
You assess whether generated sentences correctly represent four levels of plausibility based on
thematic fit.
You also assess whether the four agents are visually distinguishable in photographs, since these
sentences will be used to generate images for a human judgment experiment.
You are rigorous, precise, and catch subtle errors that might seem acceptable at first glance.
When in doubt, be strict. A false CORRECT verdict leads to an invalid experimental stimulus that
could waste a patient session. A false INCORRECT verdict only triggers one more refinement
iteration.
You return only valid JSON.

=== USER TEMPLATE ===

TASK
You will be given a verb (root form) and a set of four generated event sentences at the end of
this prompt. The sentences vary only in their AGENT (a human professional or social role) while
keeping the verb form, patient, instrument, and location identical. Your job is to judge whether
each sentence is correctly assigned to its plausibility condition, whether the four agents are
visually distinguishable and identifiable, and whether the scene itself can support the full
plausibility gradient.

---
EXPERIMENTAL CONTEXT
---
These sentences will be converted into images shown to brain injury patients who must make quick
plausible/implausible judgments. The patients see ONLY the image -- no text, no captions, no
  labels.
Therefore:
  (1) Each agent's profession must be INSTANTLY recognizable from appearance alone.
  (2) The plausibility distinction must be visually obvious and must not depend on subtle
    reasoning.
  (3) The scene must be concrete and unambiguous.
Keep this context in mind throughout your evaluation. An agent whose profession cannot be
  identified
from a photograph is a critical failure, even if the thematic fit is correct.

---
BACKGROUND: SENTENCE STRUCTURE AND THEMATIC FIT
---
Each sentence follows the structure: [Agent] [verb] [patient] [instrument] [location].

Thematic fit is how naturally a particular AGENT fits as the doer of the described action
(verb + patient + instrument + location). The four conditions form a gradient:

  plausible_easy --- plausible_hard --- implausible_hard --- implausible_easy
  (clear yes)        (unsure yes)       (unsure no)          (clear no)
  HIGH fit            MODERATE fit       LOW fit              VERY LOW fit

---
CONDITION DEFINITIONS (use these to judge each agent)
---

1) PLAUSIBLE_EASY -- "clear yes" -- HIGH thematic fit
   The agent is the prototypical, most expected professional for this action in this setting.
   The event is physically possible, reasonable, and typical.
   A human would INSTANTLY accept this as normal.
   Test: Is this THE default professional you would expect in this scene?
\end{Verbatim}
\end{tcolorbox}
\caption{Evaluator Prompt. Continued in Figure~\ref{fig:evaluator-v3.1-agent-part2}.}
\label{fig:evaluator-v3.1-agent-part1}
\end{figure*}

%% file: figures/evaluator/evaluator-v3.1-agent_part2.tex
\begin{figure*}[t]
\centering
\begin{tcolorbox}[width=\textwidth, colback=gray!5, colframe=gray!75,
                  title={Evaluator Prompt (Part 2/8)}]
\begin{Verbatim}[fontsize=\footnotesize]

   CORE-JOB TEST (mandatory for plausible_easy):
   Is [verb] [patient] part of this agent's PRIMARY professional duties -- something they are
   paid or expected to do routinely? Or is the agent merely familiar with [location]?
   Ask: "If this agent were removed from [location], would [verb] [patient] stop happening there?"
   If the answer is no, the agent's connection is to the SETTING, not the ACTION. A
   setting-prototypical agent belongs in plausible_hard, not plausible_easy.
   Example: A groundskeeper on a soccer field is setting-prototypical, but "catching
   a soccer ball" is not a groundskeeper's job -- this agent is plausible_hard at best.

2) PLAUSIBLE_HARD -- "unsure yes" -- MODERATE thematic fit
   The agent could plausibly perform this action in this setting, but it is NOT their primary
     role.
   They must be NATURALLY PRESENT or COULD NATURALLY BE PRESENT in the location.
   A human would PAUSE but ultimately accept it.
   Test: Can you easily imagine a specific, realistic scenario where this professional does this?
   CRITICAL: If the agent has no natural reason to be at the location, they belong in an
   implausible category, not here.

3) IMPLAUSIBLE_HARD -- "unsure no" -- LOW thematic fit
   The agent is from a RELATED but WRONG domain. There IS semantic overlap with the
   plausible_easy agent, but the agent does NOT belong in this specific scene.
   A human would HESITATE then reject it.

   CRITICAL -- MULTI-DIMENSIONAL OVERLAP REQUIREMENT:
   The implausible_hard agent must share overlap with the plausible_easy agent on AT LEAST TWO
   of the following five dimensions:
     (a) Occupational domain (e.g., both belong to the same broad professional field)
     (b) Typical work setting/environment (e.g., both typically work in similar environments)
     (c) Tools or instruments routinely used (e.g., both routinely use similar types of tools)
     (d) Type of patient/object acted upon (e.g., both act on similar types of objects or
       entities)
     (e) Physical actions routinely performed (e.g., both routinely perform similar physical
       actions)

   A single dimension of overlap (e.g., both use sharp tools) is NOT SUFFICIENT. If the only
   connection is through the verb itself, the agent belongs in implausible_easy, not here.
   NEGATIVE TEST: Remove the shared verb from consideration. Is there STILL a reason to associate
   this agent with the scene? If not, the overlap is too thin.

   Test: Does this agent share at least two of the above dimensions with plausible_easy
   while still being wrong for THIS specific scene?
   CRITICAL: If the agent could realistically perform this action in this setting, they are
   too plausible for this category. If they have zero or only one dimension of domain connection,
   they belong in implausible_easy.

4) IMPLAUSIBLE_EASY -- "clear no" -- VERY LOW thematic fit
   The agent is from a COMPLETELY UNRELATED domain with zero semantic connection to any
   part of the event.
   A human would INSTANTLY reject it.
   Test: Is this profession maximally distant from the scene?

---
AGENT VALIDITY
\end{Verbatim}
\end{tcolorbox}
\caption{Evaluator Prompt (continued from previous page).}
\label{fig:evaluator-v3.1-agent-part2}
\end{figure*}

%% file: figures/evaluator/evaluator-v3.1-agent_part3.tex
\begin{figure*}[t]
\centering
\begin{tcolorbox}[width=\textwidth, colback=gray!5, colframe=gray!75,
                  title={Evaluator Prompt (Part 3/8)}]
\begin{Verbatim}[fontsize=\footnotesize]
---
Valid agents include living humans identified by profession or social role.
Age-based social roles with clear visual markers are acceptable (e.g., "a schoolchild" in
school uniform, "a retiree" with gray hair and glasses) when they carry distinct role
  expectations.
Plausibility must stem from ROLE-ACTION MISMATCH (whether this role typically performs this
  action),
NOT from physical incapability. Reject agents whose implausibility depends on being physically
unable rather than role-inappropriate.
Forbidden: animals, objects, fictional beings, statues, robots, corpses, body parts,
descriptions of states, pure physical descriptors without a role.

---
EVALUATION CRITERIA
---

You will evaluate six aspects IN THIS ORDER. The order matters because later assessments
depend on earlier ones.

A) AGENT CONSTRAINTS (check first)
   A1) Is every agent a living human with an identifiable profession or social role?
   A2) Are there any forbidden agents (animals, objects, statues, robots, states, pure physical
     descriptors)?
   A3) Are the verb form, patient, instrument, and location identical across all four sentences?
   A4) Does only the agent change?
   A5) Does plausibility stem from role-action mismatch (not physical incapability)?

B) PER-CONDITION CORRECTNESS (most important)
   For each of the four conditions, judge whether the assigned agent TRULY belongs in that
   plausibility level. Consider:
   - Would this agent realistically be in this location?
   - Is this action part of (or adjacent to) their professional duties?
   - How strong is the semantic overlap with the plausible_easy agent?
   - Could this agent be confused with an adjacent condition?

   ADVERSARIAL CHECK (mandatory for every condition):
   Before issuing your verdict, you MUST articulate:
     (i)  The strongest argument for why this agent belongs ONE LEVEL HIGHER on the gradient
          (more plausible than assigned).
     (ii) The strongest argument for why this agent belongs ONE LEVEL LOWER on the gradient
          (less plausible than assigned).
   Only assign CORRECT if BOTH counter-arguments are clearly weaker than the assigned placement.
   If either counter-argument is compelling, assign BORDERLINE or INCORRECT.
   Record these counter-arguments in the output.

   For PLAUSIBLE_EASY specifically, you must also perform the TYPICALITY AUDIT:
   (i)  CORE-JOB: Is [verb] [patient] in this agent's job description? Or are they merely
        familiar with [location]? Setting-familiarity alone = plausible_hard, not PE.
   (ii) REMOVAL TEST: If this agent were removed from the scene, would the action stop?
        If no, they are not the prototypical doer.
   Record the typicality audit results in the output.

   For IMPLAUSIBLE_HARD specifically, you must also perform the OVERLAP AUDIT:
   Evaluate each of the five overlap dimensions against the plausible_easy agent:
     (a) Occupational domain: Do both agents belong to the same broad professional field?
     (b) Typical setting: Do both agents typically work in similar environments?
     (c) Tools/instruments: Do both agents routinely use similar tools?
     (d) Patient/object type: Do both agents act on similar types of objects or entities?
     (e) Physical actions: Do both agents routinely perform similar physical actions?
   Count the number of overlapping dimensions. If fewer than 2, the agent cannot be
   implausible_hard -- it should be implausible_easy.
   Also apply the NEGATIVE TEST: Remove the shared verb from consideration. Is there STILL
\end{Verbatim}
\end{tcolorbox}
\caption{Evaluator Prompt (continued from previous page).}
\label{fig:evaluator-v3.1-agent-part3}
\end{figure*}

%% file: figures/evaluator/evaluator-v3.1-agent_part4.tex
\begin{figure*}[t]
\centering
\begin{tcolorbox}[width=\textwidth, colback=gray!5, colframe=gray!75,
                  title={Evaluator Prompt (Part 4/8)}]
\begin{Verbatim}[fontsize=\footnotesize]
   a reason to associate this agent with the scene? If not, the overlap is too thin.

   Assign one of these verdicts per condition:
   - CORRECT: The agent clearly belongs in this condition.
   - BORDERLINE: The agent is debatable but not clearly wrong.
   - INCORRECT: The agent belongs in a different condition.

   If INCORRECT, specify which condition the agent actually belongs in.

C) GRADIENT ORDERING
   Is the overall ordering maintained?
   plausible_easy > plausible_hard >> implausible_hard > implausible_easy
   Are any adjacent conditions swapped or collapsed (too similar to distinguish)?

D) VISUAL EVALUATION
   These sentences will be used to generate images for a human judgment experiment.
   A participant will view each image and must be able to identify the agent's profession
   from their appearance alone. Evaluate in this order:

   D0) PROFESSION IDENTIFIABILITY (evaluate FIRST, before pairwise checks):
       For each agent, ask: "If a photograph showed ONLY this person in their typical
       professional attire against a plain white background -- no scene context, no caption,
       no other people -- could a viewer correctly guess their profession or social role?"
       A profession is identifiable if it has:
         - A dedicated uniform or standardized attire distinct from civilian clothing, OR
         - Profession-specific gear or equipment worn/carried as part of the role, OR
         - A highly distinctive physical presentation associated with the role.
       Rate each agent:
          IDENTIFIABLE: Uniform/gear makes profession obvious to a naive viewer.
         AMBIGUOUS: Could be guessed with effort but easily confused with other roles.
         UNIDENTIFIABLE: Looks like a generic civilian in everyday clothing.
       Any agent rated UNIDENTIFIABLE is a critical failure for the experiment.

   D1) Does each agent's profession have a recognizable visual identity?
       A profession has a recognizable visual identity if a naive viewer could identify it from
       appearance alone -- through a distinctive uniform, professional gear, role-specific attire,
       or iconic physical presentation. Professions that rely on generic civilian clothing
       (business casual, jeans, everyday wear) are NOT visually identifiable.

   D2) Are all four agents visually distinguishable from EACH OTHER?
       Check all six pairwise combinations:
       - plausible_easy vs plausible_hard
       - plausible_easy vs implausible_hard
       - plausible_easy vs implausible_easy
       - plausible_hard vs implausible_hard
       - plausible_hard vs implausible_easy
       - implausible_hard vs implausible_easy
       For each pair, ask: Would these two look noticeably different in a photograph based
       on their typical professional attire, gear, accessories, or physical presentation?

   D3) Flag any pair that would appear nearly identical in a photo.
       Two agents are indistinguishable if they share the same general workwear, attire type,
       or lack of distinctive professional markers. Agents from the same broad domain must
       differ in at least one strong visual cue (uniform type, headgear, carried tools, etc.).

   Assign a visual verdict:
   - ALL_DISTINCT: All six pairs are visually distinguishable AND all agents are IDENTIFIABLE.
   - MOSTLY_DISTINCT: 4-5 pairs are distinguishable, 1-2 pairs are marginal, no agent is
     UNIDENTIFIABLE.
   - PROBLEMATIC: 3+ pairs are visually indistinguishable, OR any agent is UNIDENTIFIABLE.
\end{Verbatim}
\end{tcolorbox}
\caption{Evaluator Prompt (continued from previous page).}
\label{fig:evaluator-v3.1-agent-part4}
\end{figure*}

%% file: figures/evaluator/evaluator-v3.1-agent_part5.tex
\begin{figure*}[t]
\centering
\begin{tcolorbox}[width=\textwidth, colback=gray!5, colframe=gray!75,
                  title={Evaluator Prompt (Part 5/8)}]
\begin{Verbatim}[fontsize=\footnotesize]

E) COMPREHENSIVE SCENE ASSESSMENT (evaluate LAST, after all above)
   This is the final and most holistic check. Now that you have evaluated all four agents,
   the gradient, and visual distinctiveness, assess the SCENE itself. A good scene must
   satisfy ALL of the following:

   E1) INSTRUMENT CORRECTNESS: Is the instrument appropriate for the verb?
       (e.g., you erase with an eraser, not a marker; you cut with a knife, not a spoon)

   E2) CLEAR PROTOTYPE: Does the scene have a profession for whom this action is THE core job?
       If no profession is strongly prototypical, the scene may need redesign.

   E3) IMAGEABILITY: Can the scene be captured in a single still photograph?

   E4) GRADIENT SUPPORTABILITY (most critical): Can this scene support all four plausibility
       levels with distinct, correctly-placed agents?
       This is the key question. A scene FAILS gradient supportability if:
       - The domain is so narrow that any related professional is also plausible
         (e.g., "breaking a board in a martial arts dojo" -- all combat sport professionals
         are plausible, leaving no room for implausible_hard).
       - The domain is so broad that it is hard to find an implausible_hard agent with
         genuine multi-dimensional semantic overlap (no "wait, maybe..." moment).
       - Any condition was marked INCORRECT and you cannot think of a viable replacement
         agent that would correctly fill that condition within this scene.

       To assess this, ask yourself for each INCORRECT or BORDERLINE condition:
       "Can I think of at least one profession that would correctly fill this condition
       in this scene?" If yes, the scene supports the gradient (it's an agent selection
       problem). If no, the scene itself is too narrow or too broad.

   Scene verdict:
   - GOOD: All four checks pass. The scene supports the full gradient.
   - AGENT_FIXABLE: The scene is sound but some agents are wrong. Replacing agents can fix it.
   - NEEDS_REDESIGN: The scene itself cannot support the full gradient, or the instrument is
     wrong, or there is no clear prototype. The entire scene should be redesigned.
---
EVALUATION PROCEDURE
---

STEP 1: Check all agent constraints and consistency.
STEP 2: For each condition, carefully reason about whether the agent truly belongs there.
        Compare each agent against ALL four condition definitions, not just the one it was
          assigned to.
        Ask: "Where would I place this agent if I were assigning from scratch?"
        For EACH condition, perform the ADVERSARIAL CHECK: articulate the strongest argument for
        one level higher and one level lower before issuing your verdict.
        For PLAUSIBLE_EASY, also perform the TYPICALITY AUDIT: verify core-job and removal test.
        For IMPLAUSIBLE_HARD, also perform the OVERLAP AUDIT: evaluate all five dimensions and
        apply the negative test.
STEP 3: Check the overall gradient ordering.
STEP 4: Assess visual evaluation of all four agents.
        FIRST, rate each agent's profession identifiability
          (IDENTIFIABLE/AMBIGUOUS/UNIDENTIFIABLE).
        THEN, for each agent, describe their typical professional appearance.
        THEN, check all six pairwise combinations for visual similarity.
STEP 5: Perform comprehensive scene assessment. Using your findings from Steps 2-4,
        determine whether the scene itself is the root cause of any failures.
        For each INCORRECT condition, ask: "Can I think of a profession that WOULD correctly
        fill this condition in this scene?" If not, the scene needs redesign.
STEP 6: Produce your evaluation in the OUTPUT FORMAT below.

\end{Verbatim}
\end{tcolorbox}
\caption{Evaluator Prompt (continued from previous page).}
\label{fig:evaluator-v3.1-agent-part5}
\end{figure*}

%% file: figures/evaluator/evaluator-v3.1-agent_part6.tex
\begin{figure*}[t]
\centering
\begin{tcolorbox}[width=\textwidth, colback=gray!5, colframe=gray!75,
                  title={Evaluator Prompt (Part 6/8)}]
\begin{Verbatim}[fontsize=\footnotesize]
---
OUTPUT FORMAT (STRICT)
---

Return ONLY the following JSON object. No markdown fences. No text outside the JSON.

{
  "constraint_checks": {
    "all_human_professionals": true/false,
    "consistency_maintained": true/false,
    "role_action_mismatch_valid": true/false,
    "constraint_notes": "<brief note or 'All constraints met'>"
  },
  "condition_evaluations": {
    "plausible_easy": {
      "agent": "<the agent that was assigned>",
      "verdict": "CORRECT|BORDERLINE|INCORRECT",
      "actual_condition": "<if INCORRECT, which condition this agent actually belongs in;
        otherwise same as assigned>",
      "reasoning": "<1-3 sentences explaining your judgment>",
      "counter_arguments": {
        "argument_for_higher": "<N/A for plausible_easy since it is the highest level>",
        "argument_for_lower": "<strongest argument that this agent actually belongs in
          plausible_hard or lower>"
      },
      "typicality_audit": {
        "core_job_test": "<Is [verb] [patient] part of this agent's primary job duties?
          YES/NO with brief reasoning>",
        "removal_test": "<If this agent were removed from [location], would [verb]
          [patient] stop happening? YES/NO with brief reasoning>"
      }
    },
    "plausible_hard": {
      "agent": "<the agent that was assigned>",
      "verdict": "CORRECT|BORDERLINE|INCORRECT",
      "actual_condition": "<if INCORRECT, which condition this agent actually belongs in;
        otherwise same as assigned>",
      "reasoning": "<1-3 sentences explaining your judgment>",
      "counter_arguments": {
        "argument_for_higher": "<strongest argument that this agent actually belongs in
          plausible_easy>",
        "argument_for_lower": "<strongest argument that this agent actually belongs in
          implausible_hard or lower>"
      }
    },
    "implausible_hard": {
      "agent": "<the agent that was assigned>",
      "verdict": "CORRECT|BORDERLINE|INCORRECT",
      "actual_condition": "<if INCORRECT, which condition this agent actually belongs in;
        otherwise same as assigned>",
      "reasoning": "<1-3 sentences explaining your judgment>",
      "counter_arguments": {
        "argument_for_higher": "<strongest argument that this agent actually belongs in
          plausible_hard or higher>",
        "argument_for_lower": "<strongest argument that this agent actually belongs in
          implausible_easy>"
      },
      "overlap_audit": {
        "occupational_domain": {"overlaps": true/false, "reasoning": "<brief>"},
        "typical_setting": {"overlaps": true/false, "reasoning": "<brief>"},
        "tools_instruments": {"overlaps": true/false, "reasoning": "<brief>"},
        "patient_object_type": {"overlaps": true/false, "reasoning": "<brief>"},
\end{Verbatim}
\end{tcolorbox}
\caption{Evaluator Prompt (continued from previous page).}
\label{fig:evaluator-v3.1-agent-part6}
\end{figure*}

%% file: figures/evaluator/evaluator-v3.1-agent_part7.tex
\begin{figure*}[t]
\centering
\begin{tcolorbox}[width=\textwidth, colback=gray!5, colframe=gray!75,
                  title={Evaluator Prompt (Part 7/8)}]
\begin{Verbatim}[fontsize=\footnotesize]
        "physical_actions": {"overlaps": true/false, "reasoning": "<brief>"},
        "overlap_count": <number of true dimensions>,
        "sufficient": true/false,
        "negative_test": "<Does association persist after removing the shared verb? Explain.>"
      }
    },
    "implausible_easy": {
      "agent": "<the agent that was assigned>",
      "verdict": "CORRECT|BORDERLINE|INCORRECT",
      "actual_condition": "<if INCORRECT, which condition this agent actually belongs in;
        otherwise same as assigned>",
      "reasoning": "<1-3 sentences explaining your judgment>",
      "counter_arguments": {
        "argument_for_higher": "<strongest argument that this agent actually belongs in
          implausible_hard or higher>",
        "argument_for_lower": "<N/A for implausible_easy since it is the lowest level>"
      }
    }
  },
  "gradient_ordering": {
    "correctly_ordered": true/false,
    "swapped_pairs": "<describe any swaps, or 'None'>",
    "collapsed_pairs": "<describe any conditions too similar to distinguish, or 'None'>"
  },
  "visual_evaluation": {
    "identifiability": {
      "plausible_easy": {"rating": "IDENTIFIABLE|AMBIGUOUS|UNIDENTIFIABLE", "reasoning":
        "<brief>"},
      "plausible_hard": {"rating": "IDENTIFIABLE|AMBIGUOUS|UNIDENTIFIABLE", "reasoning":
        "<brief>"},
      "implausible_hard": {"rating": "IDENTIFIABLE|AMBIGUOUS|UNIDENTIFIABLE", "reasoning":
        "<brief>"},
      "implausible_easy": {"rating": "IDENTIFIABLE|AMBIGUOUS|UNIDENTIFIABLE", "reasoning":
        "<brief>"}
    },
    "agent_appearances": {
      "plausible_easy": "<typical professional appearance of this agent>",
      "plausible_hard": "<typical professional appearance of this agent>",
      "implausible_hard": "<typical professional appearance of this agent>",
      "implausible_easy": "<typical professional appearance of this agent>"
    },
    "pairwise_checks": {
      "pe_vs_ph": {"distinguishable": true/false, "note": "<brief reason>"},
      "pe_vs_ih": {"distinguishable": true/false, "note": "<brief reason>"},
      "pe_vs_ie": {"distinguishable": true/false, "note": "<brief reason>"},
      "ph_vs_ih": {"distinguishable": true/false, "note": "<brief reason>"},
      "ph_vs_ie": {"distinguishable": true/false, "note": "<brief reason>"},
      "ih_vs_ie": {"distinguishable": true/false, "note": "<brief reason>"}
    },
    "visual_verdict": "ALL_DISTINCT|MOSTLY_DISTINCT|PROBLEMATIC",
    "indistinguishable_pairs": "<list any pairs that look too similar, or 'None'>",
    "unidentifiable_agents": "<list any agents rated UNIDENTIFIABLE, or 'None'>"
  },
  "scene_assessment": {
    "instrument_correct": true/false,
    "has_clear_prototype": true/false,
    "imageable": true/false,
    "gradient_supportable": true/false,
    "gradient_support_reasoning": "<For each INCORRECT/BORDERLINE condition, state whether you
      can think of a viable replacement agent for this scene. If you cannot, explain why the
      scene is too narrow or too broad.>",
    "scene_verdict": "GOOD|AGENT_FIXABLE|NEEDS_REDESIGN",
\end{Verbatim}
\end{tcolorbox}
\caption{Evaluator Prompt (continued from previous page).}
\label{fig:evaluator-v3.1-agent-part7}
\end{figure*}

%% file: figures/evaluator/evaluator-v3.1-agent_part8.tex
\begin{figure*}[t]
\centering
\begin{tcolorbox}[width=\textwidth, colback=gray!5, colframe=gray!75,
                  title={Evaluator Prompt (Part 8/8)}]
\begin{Verbatim}[fontsize=\footnotesize]
    "scene_notes": "<brief overall scene assessment summarizing all findings>"
  },
  "overall": {
    "score": "<number of CORRECT conditions out of 4, e.g., '3/4'>",
    "borderline_count": <number of BORDERLINE conditions>,
    "scene_verdict": "GOOD|AGENT_FIXABLE|NEEDS_REDESIGN",
    "visual_verdict": "ALL_DISTINCT|MOSTLY_DISTINCT|PROBLEMATIC",
    "summary": "<2-3 sentence overall assessment covering thematic fit, visual
      distinctiveness, identifiability, and scene quality>"
  }
}

---
INPUT TO EVALUATE (process this now)
---

Verb (root form): "{verb}"

{generated_output}

Follow the EVALUATION PROCEDURE above (Steps 1-6). Return ONLY the JSON object. No other text.
\end{Verbatim}
\end{tcolorbox}
\caption{Evaluator Prompt (continued from previous page).}
\label{fig:evaluator-v3.1-agent-part8}
\end{figure*}

%% file: appendix/D_experiment-deatils.tex
\section{Hyperparameter Details}
\label{app:hyperparameters}

This appendix lists the hyperparameters used across the
generation and evaluation runs reported in §\ref{sec:results}.

\subsection{Models}
\label{app:models}

All experiments use the model versions listed in
Table~\ref{tab:hyper_models}. Closed-source models are accessed
via provider SDKs; open-weight models are served locally with
\texttt{vLLM}.

\begin{table}[t]
\centering
\small
\resizebox{\columnwidth}{!}{%
\begin{tabular}{ll}
\toprule
\textbf{Label in paper} & \textbf{Identifier / version} \\
\midrule
GPT-5.1 & \texttt{gpt-5.1-2025-11-13} \\
Claude Sonnet~4.6 & \texttt{claude-sonnet-4-6} \\
Qwen3-Next-80B & \texttt{Qwen3-Next-80B-A3B-Instruct} \\
Qwen3-30B & \texttt{Qwen3-30B-A3B-Instruct-2507} \\
Ministral-3-14B & \texttt{Ministral-3-14B-Instruct-2512} \\
Qwen3-4B & \texttt{Qwen3-4B-Instruct-2507} \\
\midrule
Qwen3-30B-Thinking & \texttt{Qwen3-30B-A3B-Thinking-2507} \\
\bottomrule
\end{tabular}
}
\caption{Model identifiers. The six rows above the mid-rule are
used as generators throughout the paper; Qwen3-30B-Thinking is
used only as an open-source evaluator.}
\label{tab:hyper_models}
\end{table}

\subsection{Generation}
\label{app:generation}

Generators use sampling at $T = 0.3$ across all models, with
up to 16{,}384 output tokens. Native reasoning/thinking modes
are disabled on generators since they either pin temperature
to fixed values or return summarised traces; the reasoning
scratchpad introduced in §\ref{sec:generator} preserves full
reasoning at $T = 0.3$.

For R2V, the generator samples 20 candidate agents per
stimulus set with verbalized plausibility scores,
then commits one agent per condition from the tiers PE
$\in [0.80, 1.00]$, PH $\in [0.40, 0.65]$, IH
$\in [0.15, 0.35]$, and IE $=$ lowest-scoring candidate.
A fallback of 5 additional candidates is sampled if no
tier-appropriate agent passes quality checks. R2R and R2VR
run up to $k = 3$ iterations and exit early once the in-loop
evaluator returns a GOOD scene verdict with all four
per-condition verdicts marked CORRECT.

\subsection{Evaluation}
\label{app:evaluation}

GPT-5.1 evaluators run with \texttt{reasoning\_effort = high}
at the provider default temperature; Qwen3-30B-Thinking
evaluators run with $T = 0.6$ and an 8{,}000-token think
budget. Both apply the multi-step prompt described in
§\ref{sec:evaluator} that requires per-condition adversarial
counter-arguments, an overlap audit at the IH boundary, and a
holistic scene assessment. Maximum output tokens are 24{,}576
for in-loop (iterative) evaluation and 32{,}768 for standalone
(post-hoc) evaluation.

\begin{table}[t]
\centering
\small
\resizebox{\columnwidth}{!}{%
\begin{tabular}{lccc}
\toprule
\textbf{Role} & \textbf{$T$} & \textbf{Reasoning} & \textbf{Max tok.} \\
\midrule
Generator (all models) & 0.3 & disabled & 16{,}384 \\
Evaluator: GPT-5.1 & default & high effort & 24{,}576 \\
Evaluator: Qwen3-30B-Think & 0.6 & 8k budget & 32{,}768 \\
\bottomrule
\end{tabular}
}
\caption{Decoding parameters by role.}
\label{tab:hyper_decoding}
\end{table}

\subsection{Compute}
\label{app:compute}

Open-weight runs executed on a SLURM-managed GPU cluster
with NVIDIA L40S and RTX 6000 GPUs. Local models were served
with \texttt{vLLM} (\texttt{--dtype auto}, max model context
65{,}536 tokens) using tensor parallelism matched to model
size: TP$=1$ for the 4B and 14B models, TP$=2$ for the 30B
model, and TP$=4$ for the 80B model. Single-shot generation
jobs complete in 0.5--3.5 hours per model; iterative
generation runs in 12--18 hours. Closed-source API calls
require no GPU allocation.
Code, prompts, and SLURM scripts will be released upon
acceptance.

%% file: appendix/E_evaluator-scaling.tex
\section{Evaluator Scaling}
\label{sec:evaluator_scaling}

We extend the AAT validation from a single judge
(§\ref{sec:eval_validation}) to six judge models spanning
closed-source frontier (GPT-5.1, Claude Sonnet 4.6),
closed-source mid-tier (GPT-5.4-mini), and three open-source
scales (Qwen3-Next-80B-A3B, Qwen3-30B-A3B, Qwen3-4B). Each is
run in \textit{greedy} mode (no model-native reasoning) and
\textit{reasoning} mode (thinking or extended reasoning, where
available). L-H agreement is compared to the H-H baseline
($\kappa = .806$ weighted, $\kappa = .529$ unweighted), and
AAT non-inferiority is tested at $\delta = 0.05$. The
experiment, summarised in
Table~\ref{tab:evaluator_scaling}, addresses two questions:
(i) which models are recommendable as evaluators, and
(ii) whether test-time reasoning is necessary or greedy
decoding suffices.

\textbf{Reasoning helps almost everywhere.} Across 9 of 10
model-metric comparisons where both modes are tested,
reasoning beats greedy. The lone exception (Sonnet 4.6
unweighted: $.528$ vs $.521$) is sub-noise. Gains are largest
on mid-tier and small models: GPT-5.4-mini gains $+0.030$
weighted and $+0.029$ unweighted; Qwen3-4B gains $+0.028$
weighted. Test-time reasoning is therefore the default
operating mode for the evaluator.

\textbf{Only GPT-5.1 with reasoning passes both AAT criteria.}
On weighted $\kappa$ (ordering), every model where reasoning is
tested except Qwen3-4B passes AAT, so the gradient ordering is
broadly recoverable across the lineup. On unweighted $\kappa$
(exact classification), only GPT-5.1 with reasoning passes.
Exact classification is the criterion we use to recommend an
evaluator for downstream use, and it separates GPT-5.1 from the
rest of the field.

\textbf{Mid-tier open-source preserves ordering; 4B is below
the floor.} Qwen3-30B-A3B with reasoning ($\kappa = .811$
weighted) matches Sonnet 4.6 greedy ($.815$). For pipelines
that need ordering but not exact classification, e.g., as the
iteration feedback source in §\ref{sec:gen_results} rather than
the final tier-assignment evaluator, this is a viable
open-source operating point. Qwen3-4B, by contrast, fails AAT
on both metrics in both modes and is not usable as a
graded-plausibility evaluator.

\input{tables/tab_evaluators}

%% file: tables/tab_evaluators.tex
\begin{table*}[t]
\centering

\resizebox{\textwidth}{!}{%
\begin{tabular}{@{}l ccc ccc c@{}}
\toprule
& \multicolumn{3}{c}{\textbf{Weighted $\kappa$ (ordering)}} & \multicolumn{3}{c}{\textbf{Unweighted $\kappa$ (exact)}} & \\
\cmidrule(lr){2-4} \cmidrule(lr){5-7}
\textbf{Judge Model} & \textbf{L-H $\kappa$} & \textbf{Boot 5th} & \textbf{AAT $\boldsymbol{\delta{=}0.05}$} & \textbf{L-H $\kappa$} & \textbf{Boot 5th} & \textbf{AAT $\boldsymbol{\delta{=}0.05}$} & \textbf{4-pt} \\
\midrule
GPT-5.1            & .797 $\mid$ \textbf{.825} & $-$.042 $\mid$ $-$\textbf{.005} & \cmark\ $\mid$ \cmark & .488 $\mid$ \textbf{.540} & $-$.085 $\mid$ $-$\textbf{.026} & \xmark\ $\mid$ \cmark & 72 $\mid$ \textbf{82} \\
Sonnet 4.6         & .815 $\mid$ \textbf{.822} & $-$.020 $\mid$ $-$\textbf{.009} & \cmark\ $\mid$ \cmark & \textbf{.528} $\mid$ .521 & $-$\textbf{.043} $\mid$ $-$.053 & \cmark\ $\mid$ \xmark & \textbf{78} $\mid$ 77 \\
GPT-5.4-mini       & .775 $\mid$ \textbf{.805} & $-$.070 $\mid$ $-$\textbf{.031} & \xmark\ $\mid$ \cmark & .477 $\mid$ \textbf{.506} & $-$.101 $\mid$ $-$\textbf{.070} & \xmark\ $\mid$ \xmark & 74 $\mid$ \textbf{76} \\
Qwen3-30B-A3B      & .799 $\mid$ \textbf{.811} & $-$.042 $\mid$ $-$\textbf{.021} & \cmark\ $\mid$ \cmark & .511 $\mid$ \textbf{.516} & $-$.065 $\mid$ $-$\textbf{.059} & \xmark\ $\mid$ \xmark & 77 $\mid$ \textbf{80} \\
Qwen3-Next-80B-A3B & .785 $\mid$ ---           & $-$.068 $\mid$ ---             & \xmark\ $\mid$ ---    & .510 $\mid$ ---          & $-$.066 $\mid$ ---             & \xmark\ $\mid$ ---    & 77 $\mid$ --- \\
Qwen3-4B           & .735 $\mid$ \textbf{.763} & $-$.123 $\mid$ $-$\textbf{.095} & \xmark\ $\mid$ \xmark & .439 $\mid$ \textbf{.455} & $-$.137 $\mid$ $-$\textbf{.122} & \xmark\ $\mid$ \xmark & 69 $\mid$ 69 \\
\bottomrule
\end{tabular}%
}
\caption{Evaluator Scaling: Judge Model $\times$ Reasoning. Each
cell: \textit{greedy} $\mid$ \textit{reasoning}. H-H baselines:
weighted $\kappa = .806$, unweighted $\kappa = .529$. Boot 5th =
5th percentile of bootstrapped
$\Delta = \bar{\kappa}(\text{L,H}) - \bar{\kappa}(\text{H,H})$;
AAT holds when boot 5th $> -\delta$. $n = 120$ sentences, 8
annotators, $B = 10{,}000$. \textbf{Bold} = better of the
greedy/reasoning pair. Each cell: \textit{greedy/no-thinking} $\mid$
\textit{reasoning/thinking}. \cmark{} = non-inferiority
($\text{boot 5th} > -0.05$); \xmark{} = fails. ``---'' = not tested due to model loading issue.}
\label{tab:evaluator_scaling}
\end{table*}

%% file: appendix/F_log-prob.tex
\section{Probing the Evaluation Floor}
\label{sec:eval_floor}

Before committing to an LLM-as-judge, we ask whether a cheaper
information-theoretic baseline can rank the four agents using only
token probabilities from the same Qwen3-30B backbone. We then ask
whether the residual IH errors of the strongest judge can be
recovered by an ensemble over weaker ones.

\subsection{Information-Theoretic Baselines}
\label{sec:surprisal}

We score each of the four candidate agents under three progressively
stronger formulations, all on the same $n=120$ sentences (30 items
$\times$ 4 conditions). For each condition, we count how often the
top-ranked agent matches the generator-assigned label.

\paragraph{evalv5: total sentence log-probability.}
We score each candidate sentence $S = (w_1, \dots, w_T)$ by
\[
\textstyle
\log P(S) = \sum_{t=1}^{T} \log P(w_t \mid w_{<t}),
\]
and rank the four agents by $\log P(S)$. This conflates fit with the
agent's marginal pretraining frequency: a high-frequency agent
(e.g.\ \emph{doctor}) wins regardless of event fit.

\paragraph{evalv5.1: agent prior removed.}
We strip the agent span and score the post-agent tokens
$w_{a+1{:}T}$ conditioned on the agent prefix $w_{1{:}a}$:
\[
\textstyle
\log P(S \setminus \text{agent}) =
\sum_{t=a+1}^{T} \log P(w_t \mid w_{<t}).
\]
This removes prior frequency but relies on left-to-right
conditioning to propagate agent--event compatibility through
the remaining tokens.

\paragraph{Surprise - frame-first conditional.}
We restructure the prompt so the frame precedes the agent ---
\textit{``Someone [verb] [patient] [instrument] [location]. This
person is a \rule{1cm}{0.4pt}''} --- and score each candidate agent
$a$ directly at the blank,
\[
\textstyle
\log P(a \mid \text{frame}) =
\sum_{t \in a} \log P(w_t \mid \text{frame}, w_{<t}).
\]
This is the strongest distributional formulation: the full event
context conditions the agent with no prior-frequency or
ordering penalty.

\paragraph{Findings.}
Frame-first conditioning helps but does not close the gap
(Table~\ref{tab:surprisal_eval}). evalv5.2 improves 4-point accuracy
to 48\% (+7 over evalv5), with most of the gain on PE ($+$3) and IE
($+$6). PH stays pinned at 9/30 across all three variants and IH does
not improve. Even fully optimized, the best surprisal method sits ten
points below the \emph{weakest} LLM judge (Qwen3-4B greedy, 58\%) and
twenty below the strongest (GPT-5.1 with reasoning, 68\%). We read
this as structural: ``unusual but plausible'' (PH) and ``related but
wrong domain'' (IH) are constraint-satisfaction judgments, not
properties of distributional support, and no amount of conditioning
extracts them from token probabilities alone.

\subsection{Can an Ensemble Recover the Residual?}
\label{sec:ih_failure}

A natural follow-up is whether the IH errors of any single judge can
be recovered by majority vote across multiple judges. Under
Condorcet's jury theorem, independent errors above 50\% individual
accuracy should aggregate into a strictly stronger jury. We test this
on the 11 evaluator configurations from \S\ref{sec:evaluator_scaling},
split into a passing group (6 configurations meeting our IH
threshold) and a failing group (5 below it), and report per-item
difficulty, pairwise error correlation $\phi$, and a majority-vote
simulation (Table~\ref{tab:failure_analysis}).

Three results show that ensembling cannot help.
First, 15/30 IH items are unsolvable: less than 50\% of passing models
get them right, and 5 items are missed by \emph{all 11}
configurations --- a fixed LLM--human disagreement that no aggregation
can resolve. Second, the failing-group jury (33\%) is worse than its
best member (47\%); the passing-group jury also degrades by 9 points.
Error correlation is high enough ($\bar\phi = .49$ within passing,
$.27$ cross-group) that majority vote amplifies shared mistakes
rather than canceling independent ones. Third, the pass--fail accuracy
gap collapses on hard items (8\% vs.\ 13\%) after being 26 points on
easy items, indicating that capability matters only up to a ceiling
fixed by item ambiguity.

The two probes bracket the operating regime of a viable judge from
opposite sides: distributional shortcuts cannot reach the LLM floor,
and ensembling cannot lift its ceiling. Reliability claims must
therefore be made conditional on items being above the LLM--human
ambiguity threshold rather than across the full distribution.

\input{tables/tab_surprisal_eval}
\input{tables/tab_llm_judge_failure_analysis}

%% file: tables/tab_surprisal_eval.tex
\begin{table*}[t]
\centering
\small
\begin{tabular}{@{}llcccccc@{}}
\toprule
& & \multicolumn{4}{c}{\textbf{Per-condition match (/30)}} & & \\
\cmidrule(lr){3-6}
\textbf{Method} & \textbf{Signal used} & \textbf{PE} & \textbf{PH} & \textbf{IH} & \textbf{IE} & \textbf{4-pt (/120)} & \textbf{Binary (/120)} \\
\midrule
\multicolumn{8}{@{}l}{\textit{Information-theoretic evaluation (Qwen3-30B)}} \\
\midrule
evalv5    & Total sentence $\log P$                         & 15 & 9 & 12 & 13 & 49 (41\%) & 82 (68\%) \\
evalv5.1  & Frame-only $\log P$ (agent prior removed)       & 12 & 9 & 14 & 13 & 48 (40\%) & 86 (72\%) \\
Surprise  & Frame-first $P(\text{agent} \mid \text{frame})$ & 18 & 9 & 12 & 19 & 58 (48\%) & 86 (72\%) \\
\midrule
\multicolumn{8}{@{}l}{\textit{LLM-as-judge evaluation}} \\
\midrule
Worst  & Qwen3-4B, greedy      & 27 & 7 &  7 & 28 & 69 (58\%) &  98 (82\%) \\
Best   & GPT-5.1, reasoning  & 28 & 9 & 17 & 28 & 82 (68\%) & 104 (87\%) \\
\bottomrule
\end{tabular}
\caption{Information-theoretic metrics vs.\ LLM-as-judge evaluation. All surprisal methods use Qwen3-30B via sentence log-probability ranking ($n = 120$ sentences, 30 items $\times$ 4 conditions). PE identification was the primary motivation; none achieve reliable PE detection (human-aligned PE $\geq$ 27/30). PH remains flat at 9/30 across all information-theoretic methods --- token probability cannot capture ``unusual but plausible.'' IH shows no improvement --- multi-dimensional overlap reasoning requires deliberative evaluation.}
\label{tab:surprisal_eval}
\end{table*}

%% file: tables/tab_llm_judge_failure_analysis.tex
\begin{table}[t]
\centering
\small
\begin{tabular}{@{}lcc@{}}
\toprule
\textbf{Metric} & \textbf{Passing} & \textbf{Failing} \\
\midrule
Models in group           & 6         & 5 \\
IH accuracy range         & 43--57\%  & 23--47\% \\
Best individual           & 57\%      & 47\% \\
Jury (majority vote)      & 48\%      & 33\% \\
$\Delta$ jury vs best     & $-$9\%    & $-$14\% \\
Mean $\phi$ (within)      & .493      & .281 \\
Mean $\phi$ (cross-group) & \multicolumn{2}{c}{.266} \\
\midrule
\multicolumn{3}{@{}l}{\textit{Item difficulty (30 IH items)}} \\
\midrule
Easy ($>$66\% correct)    & \multicolumn{2}{c}{9 items} \\
Medium (34--66\%)         & \multicolumn{2}{c}{9 items} \\
Hard ($\leq$33\%)         & \multicolumn{2}{c}{12 items (5 at 0\%)} \\
\midrule
\multicolumn{3}{@{}l}{\textit{Pass--fail accuracy gap by difficulty}} \\
\midrule
Easy items                & 93\%  & 67\% \\
Medium items              & 54\%  & 38\% \\
Hard items                & 8\%   & 13\% \\
\midrule
\multicolumn{3}{@{}l}{\textit{Capability threshold}} \\
\midrule
Solvable ($\geq$80\% pass correct) & \multicolumn{2}{c}{10 items (gap +30\%)} \\
Ambiguous (50--79\%)       & \multicolumn{2}{c}{5 items (gap +39\%)} \\
Unsolvable ($<$50\%)       & \multicolumn{2}{c}{15 items (gap $-$5\%)} \\
\bottomrule
\end{tabular}
\caption{IH failure analysis summary. Jury = majority vote over model subset. $\phi$ = mean pairwise phi coefficient (error correlation). Condorcet assumes independent errors.}
\label{tab:failure_analysis}
\end{table}

%% file: appendix/G_verb-difficulty.tex
\section{Verb-Level Evidence for Slot-Dependent Difficulty}
\label{app:slot_dependency}

We expand the slot-dependency analysis from
§\ref{sec:slot_dependency} with three complementary views.

\paragraph{Per-verb GOLD distribution
(Table~\ref{tab:verb_difficulty}).}
Aggregating across all models and methods in agent-varying
generation, per-verb GOLD rates span $42.4\%$ (\textit{throw})
to $3.4\%$ (\textit{tickle}, \textit{pay}). The distribution
divides into three bands: Easy ($\geq 20\%$ GOLD, $n=11$),
Moderate (10--19\%, $n=34$), and Hard ($<10\%$, $n=15$). The
Hard band concentrates two kinds of verbs: (a) actions tied to
no specific occupation (\textit{eat}, \textit{drink},
\textit{tickle}, \textit{pay}), which leave no prototypical PE
agent, and (b) actions with very narrow occupational scope
(\textit{tow}, \textit{iron}, \textit{erase}), which leave no
room for a distinct IH agent. Both failure modes are properties
of the agent slot, not of the action itself.

\paragraph{Aggregate slot comparison
(Table~\ref{tab:slot_aggregate}).}
With the patient slot varied, mean R2V GOLD rises from $11.3\%$
(agent) to $16.4\%$ (patient) across the five models with
both-slot coverage. The gain concentrates where agent-varying
was weakest: Qwen3-Next-80B-A3B improves from $6.7\%$ to
$21.7\%$. RB shows a smaller and reverse aggregate shift
($10.0\%$ agent, $6.7\%$ patient), driven by Sonnet 4.6's
outlier agent-side performance ($28.3\%$); without iteration,
neither slot can be reliably exploited.

\paragraph{Per-verb slot overlap
(Figure~\ref{fig:slot_perverb}).}
At the per-(model, verb) level, only $3\%$ of pairs achieve
GOLD in both slot conditions under R2V; the remaining GOLD
pairs split between agent-only ($8\%$) and patient-only
($13\%$). Verbs that succeed in agent-varying generation are
largely a different set from those that succeed in
patient-varying generation. A single-slot ceiling on GOLD rate
is a ceiling on what that slot can support for the verb, not a
ceiling on what the framework can generate; comprehensive
stimulus coverage requires varying multiple slots.

\input{tables/tab_verb_difficulty}

\input{figures/slot_perverb}

%% file: tables/tab_verb_difficulty.tex
\begin{table}[t]
\centering
\resizebox{\columnwidth}{!}{%
\begin{tabular}{l r r r r r l}
\toprule
\textbf{Verb} & \textbf{GOLD} & \textbf{SILVER} & \textbf{FAIL} & \textbf{$n$} & \textbf{GOLD\%} & \textbf{Difficulty} \\
\midrule
\textit{catch} & 26 & 11 & 15 & 56 & 46.4 & Easy \\
\textit{throw} & 25 & 7 & 22 & 59 & 42.4 & Easy \\
\textit{fish} & 20 & 4 & 28 & 59 & 33.9 & Easy \\
\textit{shoot} & 19 & 5 & 28 & 58 & 32.8 & Easy \\
\textit{fight} & 16 & 1 & 40 & 58 & 27.6 & Easy \\
\textit{saw} & 14 & 5 & 37 & 58 & 24.1 & Easy \\
\textit{thread} & 11 & 3 & 40 & 54 & 20.4 & Easy \\
\textit{water} & 12 & 4 & 42 & 59 & 20.3 & Easy \\
\textit{sweep} & 12 & 3 & 44 & 59 & 20.3 & Easy \\
\textit{pour} & 12 & 7 & 39 & 59 & 20.3 & Easy \\
\textit{feed} & 12 & 9 & 38 & 59 & 20.3 & Easy \\
\addlinespace[3pt]
\textit{sew} & 11 & 5 & 40 & 58 & 19.0 & Moderate \\
\textit{shave} & 11 & 6 & 39 & 58 & 19.0 & Moderate \\
\textit{grate} & 11 & 2 & 45 & 59 & 18.6 & Moderate \\
\textit{build} & 11 & 9 & 38 & 59 & 18.6 & Moderate \\
\textit{burn} & 11 & 3 & 41 & 59 & 18.6 & Moderate \\
\textit{measure} & 11 & 4 & 41 & 59 & 18.6 & Moderate \\
\textit{rake} & 11 & 2 & 46 & 59 & 18.6 & Moderate \\
\textit{stir} & 10 & 2 & 45 & 59 & 16.9 & Moderate \\
\textit{spread} & 10 & 4 & 45 & 59 & 16.9 & Moderate \\
\textit{comb} & 10 & 6 & 41 & 59 & 16.9 & Moderate \\
\textit{break} & 10 & 5 & 40 & 59 & 16.9 & Moderate \\
\textit{melt} & 10 & 6 & 41 & 59 & 16.9 & Moderate \\
\textit{touch} & 9 & 10 & 37 & 58 & 15.5 & Moderate \\
\textit{paint} & 9 & 7 & 42 & 58 & 15.5 & Moderate \\
\textit{conduct} & 9 & 5 & 38 & 58 & 15.5 & Moderate \\
\textit{dig} & 9 & 7 & 38 & 59 & 15.3 & Moderate \\
\textit{light} & 9 & 2 & 44 & 59 & 15.3 & Moderate \\
\textit{wash} & 9 & 6 & 42 & 59 & 15.3 & Moderate \\
\textit{peel} & 9 & 1 & 49 & 59 & 15.3 & Moderate \\
\textit{weigh} & 9 & 11 & 34 & 59 & 15.3 & Moderate \\
\textit{cook} & 8 & 5 & 43 & 59 & 13.6 & Moderate \\
\textit{cut} & 8 & 3 & 48 & 59 & 13.6 & Moderate \\
\textit{write} & 8 & 6 & 45 & 61 & 13.1 & Moderate \\
\textit{scoop} & 7 & 4 & 45 & 58 & 12.1 & Moderate \\
\textit{curl} & 7 & 5 & 44 & 58 & 12.1 & Moderate \\
\textit{brush} & 7 & 7 & 40 & 58 & 12.1 & Moderate \\
\textit{dust} & 7 & 1 & 50 & 59 & 11.9 & Moderate \\
\textit{drill} & 7 & 7 & 44 & 59 & 11.9 & Moderate \\
\textit{glue} & 7 & 10 & 37 & 59 & 11.9 & Moderate \\
\textit{swat} & 7 & 4 & 48 & 59 & 11.9 & Moderate \\
\textit{vacuum} & 7 & 5 & 47 & 59 & 11.9 & Moderate \\
\textit{open} & 7 & 5 & 45 & 59 & 11.9 & Moderate \\
\textit{plant} & 7 & 6 & 45 & 59 & 11.9 & Moderate \\
\textit{watch} & 7 & 1 & 48 & 59 & 11.9 & Moderate \\
\addlinespace[3pt]
\textit{pinch} & 5 & 6 & 39 & 55 & 9.1 & Hard \\
\textit{weave} & 5 & 10 & 34 & 58 & 8.6 & Hard \\
\textit{carry} & 5 & 6 & 44 & 59 & 8.5 & Hard \\
\textit{look} & 5 & 8 & 42 & 59 & 8.5 & Hard \\
\textit{draw} & 5 & 3 & 49 & 59 & 8.5 & Hard \\
\textit{drink} & 4 & 1 & 51 & 59 & 6.8 & Hard \\
\textit{bury} & 4 & 8 & 40 & 59 & 6.8 & Hard \\
\textit{erase} & 4 & 2 & 51 & 59 & 6.8 & Hard \\
\textit{type} & 3 & 4 & 51 & 59 & 5.1 & Hard \\
\textit{iron} & 3 & 2 & 53 & 59 & 5.1 & Hard \\
\textit{tow} & 3 & 21 & 32 & 59 & 5.1 & Hard \\
\textit{eat} & 3 & 1 & 56 & 61 & 4.9 & Hard \\
\textit{knit} & 2 & 5 & 48 & 56 & 3.6 & Hard \\
\textit{pay} & 2 & 2 & 52 & 58 & 3.4 & Hard \\
\textit{tickle} & 2 & 5 & 50 & 59 & 3.4 & Hard \\
\bottomrule
\end{tabular}
}
\caption{Per-verb GOLD rate across all models and methods (60 verbs, sorted by GOLD\%). Difficulty: Easy ($\geq$20\%), Moderate (10--19\%), Hard ($<$10\%).}
\label{tab:verb_difficulty}
\end{table}

%% file: figures/slot_perverb.tex
\begin{figure*}[t]
\centering
\includegraphics[width=\textwidth]{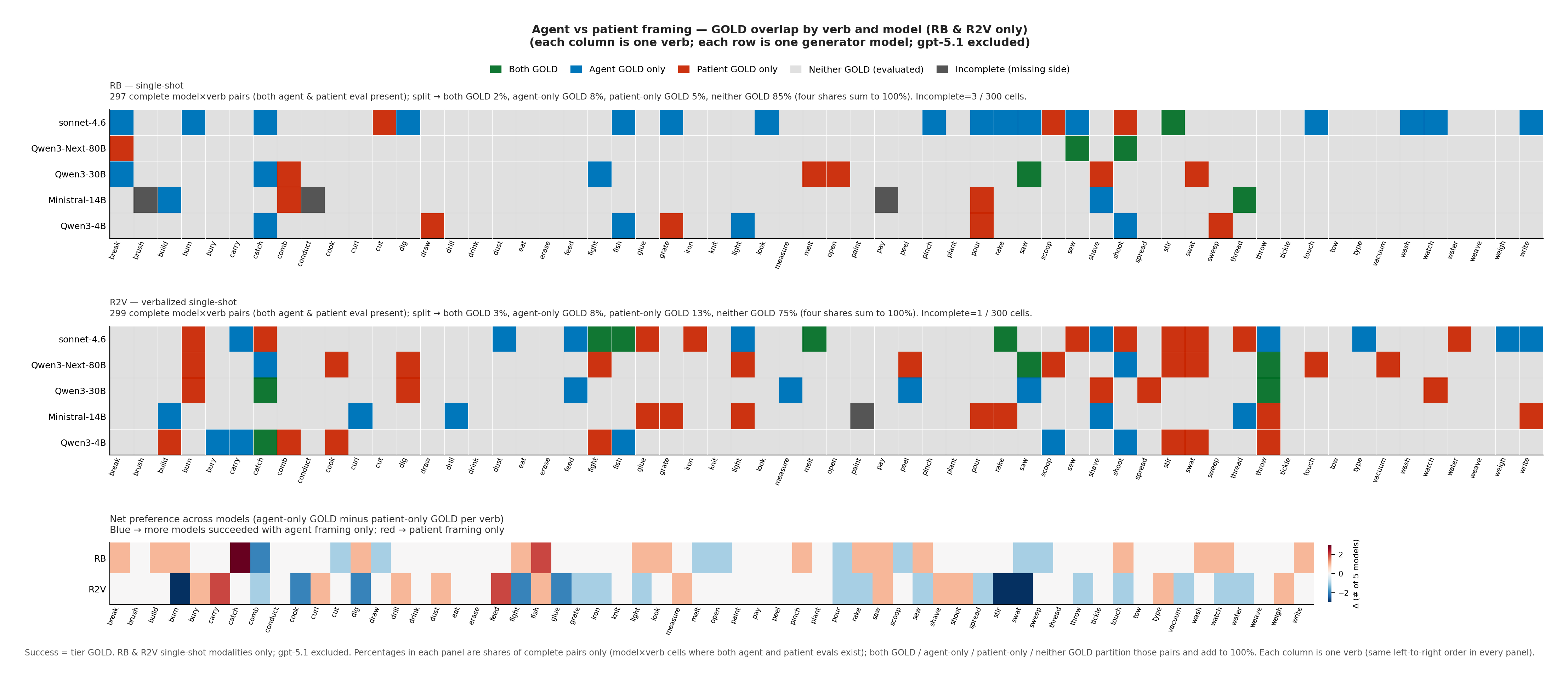}
\caption{Per-verb GOLD outcome under agent vs.\ patient slot
variation, for each model. Each cell is one (model, verb) pair
where both slot conditions were evaluated; cell color indicates
which slot(s) achieved GOLD. The both-GOLD share is $2\%$
(RB, top) and $3\%$ (R2V, bottom), confirming that the
agent-GOLD and patient-GOLD verb sets are largely disjoint.}
\label{fig:slot_perverb}
\end{figure*}

%% file: appendix/H_token-cost.tex
\section{Output Token Costs}
\label{app:token_costs}

Table~\ref{tab:token_costs} reports the mean per-verb generator
output tokens for each method across all six generators, computed
over the 60-verb set. Per-verb counts sum all generator API call
outputs in a run; for R2R and R2VR, this includes outputs across
all refinement iterations but excludes the evaluator's tokens.
Iterative methods average 2.4--2.8 rounds per verb, with early
exit when the evaluator returns a GOOD scene verdict and four
CORRECT condition verdicts (§\ref{sec:experiments}).

\input{tables/tab_token_cost}

\paragraph{Iteration cost.}
On the closed-source generators where R2R was run, total output
tokens scale by 2.72$\times$ (GPT-5.1) to 3.51$\times$ (Sonnet~4.6)
relative to RB at matched reasoning configuration; R2VR scales by
2.02$\times$ to 2.25$\times$ relative to R2V. The mean
per-iteration generator output (total $\div$ rounds) is
comparable to the corresponding single-shot method
(e.g., GPT-5.1 R2R averages 2{,}634 tokens per iteration vs
RB's 2{,}633), indicating that the cost inflation in iterative
methods comes from running multiple rounds rather than heavier
per-call generation.

\paragraph{Reasoning scratchpad cost.}
Within a method, the global reasoning scratchpad adds
$+1{,}690$ tokens to RB on average (3.74$\times$ the no-scratchpad
baseline) and $+1{,}856$ tokens to R2V (1.88$\times$). The larger
relative increase for RB reflects that no-scratchpad RB is also
the shortest method (mean 557--685 output tokens across
generators), so the added reasoning trace dominates total length.

%% file: tables/tab_token_cost.tex
\begin{table}[t]
\centering
\small
\begin{tabular}{l rrrr}
\toprule
\textbf{Model} & \textbf{RB} & \textbf{R2V} & \textbf{R2R} & \textbf{R2VR} \\
\midrule
GPT-5.1            & 2{,}633 & 4{,}819 & 7{,}161  & 9{,}759 \\
Sonnet~4.6         & 4{,}421 & 6{,}795 & 15{,}531 & 15{,}276 \\
Qwen3-Next-80B     & 1{,}645 & 3{,}083 & ---      & --- \\
Qwen3-30B          & 1{,}659 & 2{,}982 & ---      & --- \\
Ministral-14B      & 1{,}663 & 2{,}900 & ---      & --- \\
Qwen3-4B           & 1{,}821 & 3{,}159 & ---      & --- \\
\bottomrule
\end{tabular}
\caption{Mean generator output tokens per verb on the 60-verb
set. R2R/R2VR sum tokens across all refinement iterations
(mean 2.4--2.8 rounds per verb). Open-source iterative runs
used the Qwen judge for both iteration feedback and evaluation
(§\ref{sec:experiments}) and are omitted here to preserve
like-for-like comparisons against the GPT-judged closed-source
pipeline.}
\label{tab:token_costs}
\end{table}

%% file: appendix/I_concordance.tex
\section{Cross-Evaluator GOLD Concordance}
\label{app:concordance}

Whether an open-source pipeline can substitute for the
closed-source one hinges on a directional question: when the
Qwen3-30B judge flags a stimulus set as \textsc{Gold}, does the
GPT-5.1 judge also flag it as \textsc{Gold}? We answer this on
the closed-source iterative subset---R2R and R2VR on GPT-5.1 and
Sonnet~4.6, $N=235$---the configuration that produces the
paper's primary high-quality results (§\ref{sec:gen_results}).
Single-shot methods (RB, R2V) are excluded because \textsc{Gold}
is rare under either judge in those runs, and the small
\textsc{Gold} pool inflates variance.

Of the 235 cells, both judges agreed on \textsc{Gold} in 97 and
on non-\textsc{Gold} in 35. The remaining 103 split asymmetrically:
GPT alone flagged \textsc{Gold} in 81 cells, Qwen alone in 22
(Table~\ref{tab:concordance_2x2}).

\begin{table}[h]
\centering
\small
\begin{tabular}{lcc}
\toprule
                 & Qwen $\neq$ \textsc{Gold} & Qwen $=$ \textsc{Gold} \\
\midrule
GPT $\neq$ \textsc{Gold} & 35 & 22 \\
GPT $=$ \textsc{Gold}    & 81 & 97 \\
\bottomrule
\end{tabular}
\caption{\textsc{Gold}-tier contingency on the closed-source
iterative subset ($N=235$). Rows: GPT-5.1 judge; columns:
Qwen3-30B-Thinking judge.}
\label{tab:concordance_2x2}
\end{table}

\paragraph{Primary statistic.}
Conditional on Qwen flagging \textsc{Gold}, GPT also flags
\textsc{Gold} in $97/119 = 81.5\%$ of cells (Wilson 95\% CI:
$[73.6\%, 87.5\%]$). The CI excludes 50\%, ruling out chance
agreement. The reverse conditional is lower:
$P(\text{Qwen}\!=\!\textsc{Gold} \mid \text{GPT}\!=\!\textsc{Gold})
= 97/178 = 54.5\%$. Qwen is therefore the stricter judge, and
its \textsc{Gold} set is a higher-precision subset of GPT's.
Restricting downstream stimulus selection to Qwen-flagged
\textsc{Gold} yields a smaller pool, but one that GPT also
endorses with high probability.

\paragraph{Consistency across iterative methods.}
The two iterative methods show comparable concordance:
$P(\text{GPT}\mid\text{Qwen}) = 80.4\%$ on R2R ($N=115$) and
$82.5\%$ on R2VR ($N=120$). Figure~\ref{fig:judge-verb-heatmap}
shows the per-verb structure of agreement and disagreement; the
\textsc{Gold}-overlap pattern (green cells) is distributed
across verbs rather than concentrated in a small subset,
confirming that the 81.5\% is a population-level property of the
two judges rather than an artefact of a few easy verbs.

\input{figures/fig_judge_concordance}

%% file: figures/fig_judge_concordance.tex
\begin{figure*}[h]
\centering
\includegraphics[width=\linewidth]{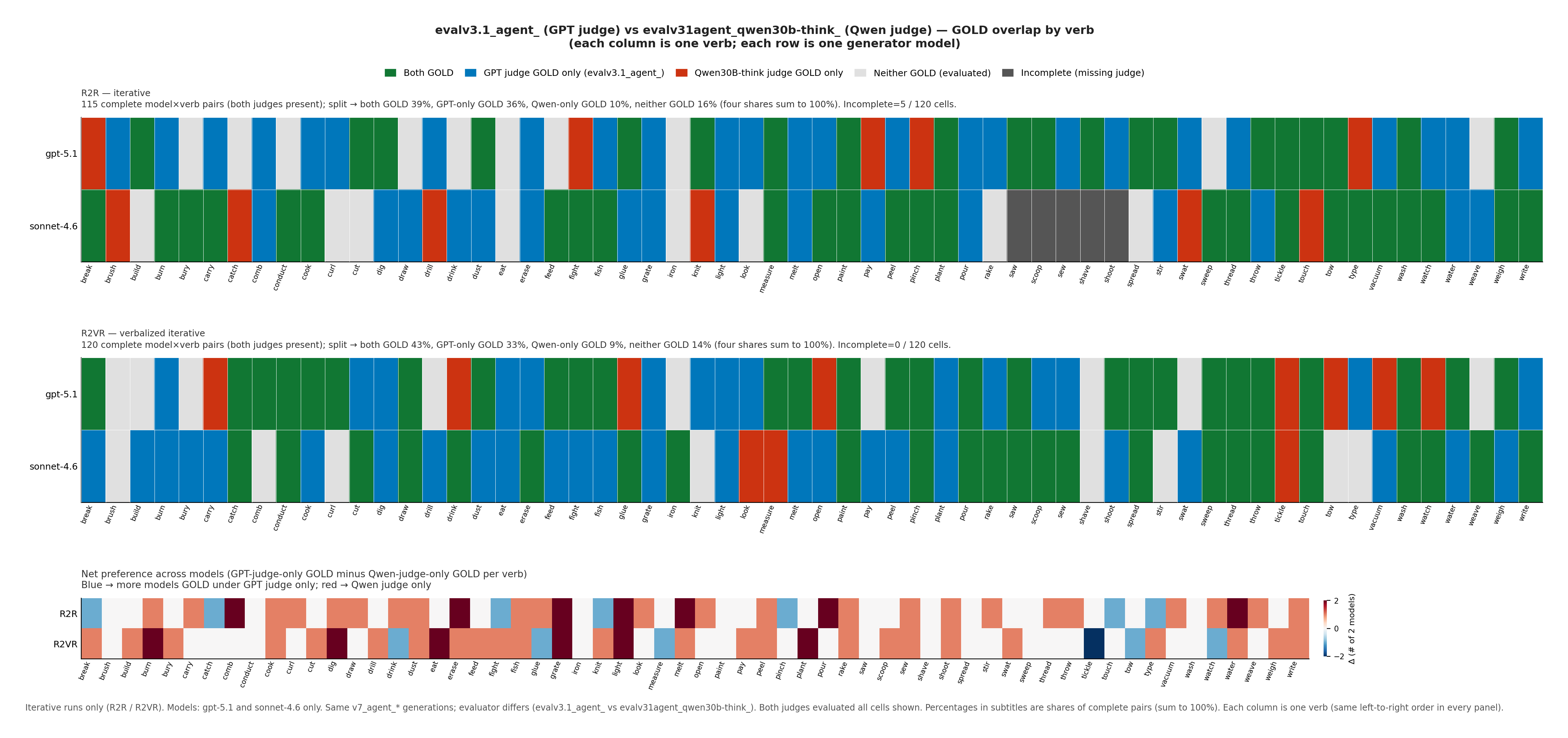}
\caption{Per-verb \textsc{Gold} overlap between the GPT-5.1 and
Qwen3-30B-Thinking judges on the closed-source iterative subset.
Top: R2R (115 complete cells). Middle: R2VR (120 complete
cells). Bottom: net preference per verb (GPT-only minus
Qwen-only count across the two generators). Green: both judges
\textsc{Gold}; blue: GPT only; red: Qwen only; light gray:
neither; dark gray: missing-judge cells.}
\label{fig:judge-verb-heatmap}
\end{figure*}

%% file: appendix/K_plausibility_gate.tex
\section{Plausibility-Gate Pass Rates}
\label{app:plausibility_gate}

The plausibility-gate pass rate is the proportion of sets for which
all four condition verdicts are correct, irrespective of scene and visual
verdicts. Under the hierarchical rule in Table~\ref{tab:tiers}, this is the
share that clears the plausibility prerequisite for a non-FAIL tier before
scene and visual verdicts determine the final quality tier.
Table~\ref{tab:plausibility_gate} compares this rate with GOLD under the GPT
judge.

\begin{table}[h]
\centering
\small
\resizebox{\columnwidth}{!}{%
\begin{tabular}{llrrrr}
\toprule
\textbf{Generator} & \textbf{Rate} & \textbf{RB} & \textbf{R2V} &
\textbf{R2R} & \textbf{R2VR} \\
\midrule
GPT-5.1    & GOLD              & 28.3 & 13.3 & 75.0 & 71.7 \\
           & Plausibility gate & 51.7 & 36.7 & 86.7 & 83.3 \\
Sonnet~4.6 & GOLD              & 28.3 & 21.7 & 74.5 & 81.7 \\
           & Plausibility gate & 43.3 & 45.0 & 89.1 & 93.3 \\
\bottomrule
\end{tabular}}
\caption{GOLD and plausibility-gate pass rates (\%) under the GPT
judge. The refinement gains persist before applying the scene and visual
gates.}
\label{tab:plausibility_gate}
\end{table}

%% file: appendix/D_appendix_annotation.tex
\section{Human Annotation Study Details}
\label{app:annotation}

\paragraph{Stimulus sampling.}
Thirty stimulus sets were sampled from the full evaluation pool via
stratified sampling across quality tiers (GOLD, SILVER, BRONZE+FAIL),
ensuring coverage of the full quality range rather than selecting
only successful outputs.

\paragraph{Survey instrument.}
Each annotator rated all 120 sentences (30 sets $\times$ 4 conditions)
on a four-point plausibility scale:
(1)~\textit{Clearly Plausible},
(2)~\textit{Somewhat Plausible},
(3)~\textit{Somewhat Implausible},
(4)~\textit{Clearly Implausible},
plus a \textit{Cannot Decide} option.
Annotators additionally rated decision difficulty per sentence
(1~=~Easy, 2~=~Moderate, 3~=~Hard).
At the set level, annotators rated the visual identifiability of
each agent (Yes / Maybe / No) and flagged any agent pairs they
judged visually indistinguishable.
No annotator selected \textit{Cannot Decide} for any item, confirming
that all stimuli were interpretable.
Instructions to annotators and sample questions are shown in Figure~\ref{fig:surveypage}

\paragraph{Gradient separability.}
For each sentence, we average the eight plausibility ratings and treat each matched four-sentence set as the repeated-measures unit ($n=30$). Because the original ratings are ordinal and every set contains all four conditions, we use a Friedman test \citep{friedman1937ranks} rather than a parametric or independent-samples test to assess whether ratings differ overall. We then compare the three adjacent pairs using paired Wilcoxon signed-rank tests \citep{wilcoxon1945individual} with Holm correction \citep{holm1979simple} for multiple comparisons. Mean ratings increase from PE 1.21 to PH 1.89, IH 2.58, and IE 3.86. The Friedman test finds an overall condition effect ($\chi^2(3)=76.5$, $p<10^{-15}$). All adjacent comparisons remain significant after Holm correction: PE--PH ($p=3.3\times10^{-4}$), PH--IH ($p=5.3\times10^{-4}$), and IH--IE ($p=5.1\times10^{-6}$).

\paragraph{Binary mapping.}
A supplementary binary analysis maps ratings 1--2 to
\textsc{plausible} and 3--4 to \textsc{implausible}, reflecting the
patient-facing judgment task in which participants make a binary
decision.
Per-condition agreement is additionally reported using Gwet's AC1
to guard against the $\kappa$ prevalence paradox at the extreme
conditions (PE and IE).

\paragraph{Annotator details.}
Eight trained annotators participated; all were fluent English
speakers familiar with psycholinguistic experimental design.
Annotation was conducted via an online survey platform.
Median completion time was 169 minutes (5.6 minutes per set).

\input{figures/fig_survey_page}

%% file: figures/fig_survey_page.tex
\begin{figure*}[t]
  \centering
  \includegraphics[width=\textwidth]{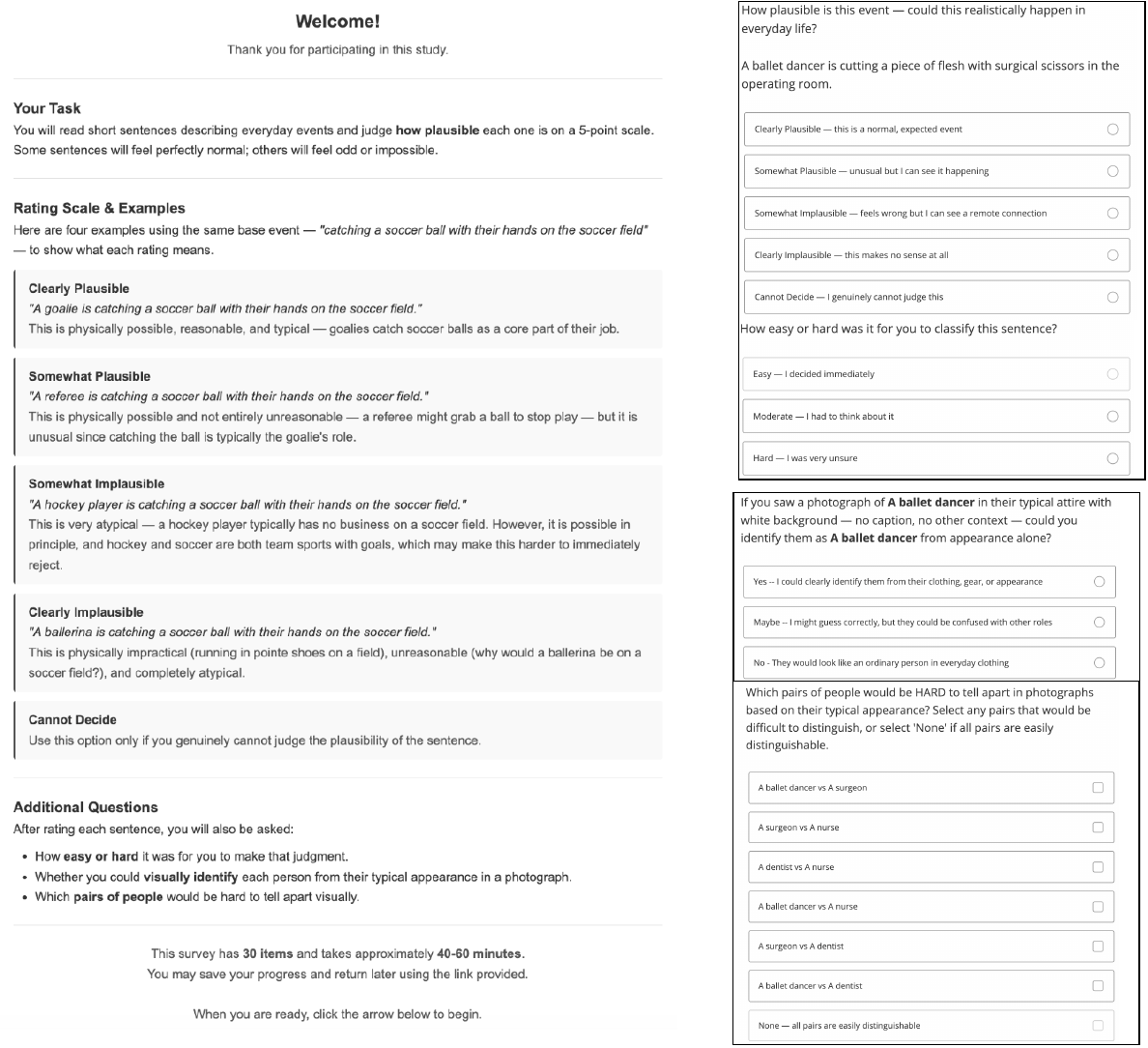}
  \caption{Survey welcome page with instructions and sample questions for per-sentence plausibility classification and visual distinctiveness}
  \label{fig:surveypage}
\end{figure*}

%% file: appendix/E_appendix_aat.tex
\section{Extended AAT Results}
\label{app:aat_detail}

\paragraph{H-H baseline.}
Krippendorff's $\alpha = 0.805$ across all 8 annotators and 120
sentences confirms strong inter-annotator agreement and that the
stimuli are interpretable (zero \textit{Cannot Decide} responses).
Mean pairwise weighted $\kappa_w = 0.806$ ($\pm 0.061$).

\paragraph{Per-condition exact match (4-point).}
Exact match between LLM and human majority:
PE = 90\%, PH = 30\%, IH = 53\%, IE = 93\%.
The low PH match reflects genuine ambiguity: 75\% of PH sentences
received a \textsc{plausible} human majority, but only 30\% matched
the LLM's assigned category exactly.
94\% of all 4-point disagreements are one-step adjacent on the
gradient; mean disagreement difficulty exceeds mean agreement
difficulty (Mann-Whitney $p < 0.001$), confirming that
disagreements cluster on items that human annotators themselves
found hard.

\paragraph{Per-condition binary AAT (AC1).}
\begin{center}
\small
\begin{tabular}{lrrrr}
\toprule
\textbf{Condition} & \textbf{Prev.} & \textbf{H-H AC1} & \textbf{L-H AC1} & \textbf{$\Delta$} \\
\midrule
PE & 93\% & 0.922 & 0.922 & $+$0.000 \\
PH & 75\% & 0.582 & 0.664 & $+$0.081 \\
IH & 43\% & 0.426 & 0.525 & $+$0.099 \\
IE &  4\% & 0.917 & 0.958 & $+$0.041 \\
\bottomrule
\end{tabular}
\end{center}
All four conditions pass non-inferiority at $\delta \leq 0.05$
(AC1 bootstrap). LLM agreement exceeds H-H at PH and IH,
the two conditions where unaided human intuition is least reliable.

\paragraph{Supporting weighted $\kappa_w$ and binary AAT.}
4-point weighted $\kappa_w$: H-H = 0.806, L-H = 0.821,
$\Delta = {+}0.015$, passing at $\delta \leq 0.01$.
Binary $\kappa$: H-H = 0.650, L-H = 0.686,
$\Delta = {+}0.036$, passing at $\delta \leq 0.01$.

\paragraph{F1 identifiability detail.}
Human identifiability distribution: Yes = 52.4\%, Maybe = 35.9\%,
No = 11.7\%.
LLM distribution: IDENTIFIABLE = 75\%, AMBIGUOUS = 22.5\%,
UNIDENTIFIABLE = 2.5\%.
F1 $\kappa$: H-H = 0.277, L-H = 0.210, $\Delta = {-}0.067$,
passing at $\delta \leq 0.15$.
The LLM is systematically more optimistic about identifiability
than human annotators, underrating the Maybe and No categories.

%% file: appendix/v7_agent_paper_examples_companion.tex
\input{tables/v7_agent_paper_examples_table}

%% file: tables/v7_agent_paper_examples_table.tex
\definecolor{tierGold}{HTML}{BA7517}
\definecolor{tierSilver}{HTML}{888780}
\definecolor{tierBronze}{HTML}{F0997B}
\definecolor{tierFail}{HTML}{E8E6DF}
\begin{table*}[p]
\centering
\scriptsize
\setlength{\tabcolsep}{2.5pt}
\begin{tabularx}{\textwidth}{l l l X X X X}
\toprule
Model & Method & Verb & PE sentence & PH sentence & IH sentence & IE sentence \\
\midrule
\rowcolor{tierGold!12}
sonnet-4.6 & RB & fish & A fly fisherman is fishing for a trout with a fishing rod on the riverbank. & A park ranger is fishing for a trout with a fishing rod on the riverbank. & A scuba diver is fishing for a trout with a fishing rod on the riverbank. & A ballet dancer is fishing for a trout with a fishing rod on the riverbank. \\
\rowcolor{tierGold!12}
sonnet-4.6 & R2V & fish & A fly fisherman is fishing a trout with a fishing rod in the river. & A park ranger is fishing a trout with a fishing rod in the river. & A scuba diver is fishing a trout with a fishing rod in the river. & A jockey is fishing a trout with a fishing rod in the river. \\
\rowcolor{tierGold!12}
sonnet-4.6 & R2R & fish & A fisherman is fishing for trout with a fishing rod at the riverbank. & A park ranger is fishing for trout with a fishing rod at the riverbank. & A scuba diver is fishing for trout with a fishing rod at the riverbank. & A firefighter is fishing for trout with a fishing rod at the riverbank. \\
\rowcolor{tierGold!12}
sonnet-4.6 & R2VR & fish & A fly fisherman is fishing a trout with a fishing rod in the river. & A park ranger is fishing a trout with a fishing rod in the river. & A scuba diver is fishing a trout with a fishing rod in the river. & A ballet dancer is fishing a trout with a fishing rod in the river. \\
\rowcolor{tierGold!12}
gpt-5.1 & RB & wash & A car wash worker is washing a car with a high-pressure hose at the car wash. & A mechanic is washing a car with a high-pressure hose at the car wash. & A firefighter is washing a car with a high-pressure hose at the car wash. & A ballet dancer is washing a car with a high-pressure hose at the car wash. \\
\rowcolor{tierGold!12}
gpt-5.1 & R2V & wash & A hairdresser is washing a customer's hair with a shower sprayer at the salon sink. & A manicurist is washing a customer's hair with a shower sprayer at the salon sink. & A dermatologist is washing a customer's hair with a shower sprayer at the salon sink. & An astronaut is washing a customer's hair with a shower sprayer at the salon sink. \\
\rowcolor{tierSilver!24}
sonnet-4.6 & R2V & dig & A construction worker is digging a trench with a shovel at a construction site. & A plumber is digging a trench with a shovel at a construction site. & An archaeologist is digging a trench with a shovel at a construction site. & A surgeon is digging a trench with a shovel at a construction site. \\
\rowcolor{tierSilver!24}
Qwen3-80B & R2V & dig & A construction worker is digging a hole with a shovel on the construction site. & A gardener is digging a hole with a shovel on the construction site. & A paleontologist is digging a hole with a shovel on the construction site. & A chef is digging a hole with a shovel on the construction site. \\
\rowcolor{tierSilver!24}
Qwen3-30B & R2V & dig & A construction worker is digging a trench with a shovel in the construction site. & A construction foreman is digging a trench with a shovel in the construction site. & An archaeologist is digging a trench with a shovel in the construction site. & A ballet dancer is digging a trench with a shovel in the construction site. \\
\rowcolor{tierSilver!24}
gpt-5.1 & RB & tow & A tow truck driver is towing a broken-down car with a tow truck on the city street. & A police officer is towing a broken-down car with a tow truck on the city street. & A garbage collector is towing a broken-down car with a tow truck on the city street. & A ballet dancer is towing a broken-down car with a tow truck on the city street. \\
\rowcolor{tierSilver!24}
sonnet-4.6 & RB & tow & A tow truck driver is towing a broken-down car with a tow truck on the highway. & A police officer is towing a broken-down car with a tow truck on the highway. & A construction worker is towing a broken-down car with a tow truck on the highway. & A chef is towing a broken-down car with a tow truck on the highway. \\
\rowcolor{tierSilver!24}
Ministral-14B & RB & tow & A tow truck driver is towing a broken-down car with a tow truck on a highway. & A police officer is towing a broken-down car with a tow truck on a highway. & A construction worker is towing a broken-down car with a tow truck on a highway. & A florist is towing a broken-down car with a tow truck on a highway. \\
\rowcolor{tierBronze!24}
Qwen3-80B & RB & carry & A construction worker is carrying a heavy bag of cement with his arms on the construction site. & A foreman is carrying a heavy bag of cement with his arms on the construction site. & A firefighter is carrying a heavy bag of cement with his arms on the construction site. & A librarian is carrying a heavy bag of cement with his arms on the construction site. \\
\rowcolor{tierBronze!24}
Qwen3-80B & RB & dig & A construction worker is digging a hole with a shovel at the construction site. & A landscaper is digging a hole with a shovel at the construction site. & A grave digger is digging a hole with a shovel at the construction site. & A chef is digging a hole with a shovel at the construction site. \\
\rowcolor{tierBronze!24}
Qwen3-80B & RB & glue & A carpenter is gluing a broken wooden chair with wood glue in a carpentry workshop. & A handyman is gluing a broken wooden chair with wood glue in a carpentry workshop. & A bookbinder is gluing a broken wooden chair with wood glue in a carpentry workshop. & A chef is gluing a broken wooden chair with wood glue in a carpentry workshop. \\
\rowcolor{tierBronze!24}
Qwen3-80B & RB & saw & A carpenter is sawing a wooden plank with a handsaw in a carpenter's workshop. & A DIY homeowner is sawing a wooden plank with a handsaw in a carpenter's workshop. & A lumberjack is sawing a wooden plank with a handsaw in a carpenter's workshop. & A chef is sawing a wooden plank with a handsaw in a carpenter's workshop. \\
\rowcolor{tierFail!18}
gpt-5.1 & RB & iron & A hotel housekeeper is ironing a dress shirt with a steam iron on an ironing board in a laundry room. & A hotel manager is ironing a dress shirt with a steam iron on an ironing board in a laundry room. & A janitor is ironing a dress shirt with a steam iron on an ironing board in a laundry room. & A firefighter is ironing a dress shirt with a steam iron on an ironing board in a laundry room. \\
\rowcolor{tierFail!18}
gpt-5.1 & R2V & iron & A hotel housekeeper is ironing a dress shirt with an electric iron on an ironing board in a laundry room. & A clothing store clerk is ironing a dress shirt with an electric iron on an ironing board in a laundry room. & A chef is ironing a dress shirt with an electric iron on an ironing board in a laundry room. & An astronaut is ironing a dress shirt with an electric iron on an ironing board in a laundry room. \\
\rowcolor{tierFail!18}
gpt-5.1 & R2R & iron & A hotel housekeeper is ironing a dress shirt with an electric iron on an ironing board in a laundry room. & A chef is ironing a dress shirt with an electric iron on an ironing board in a laundry room. & A hotel manager is ironing a dress shirt with an electric iron on an ironing board in a laundry room. & A scuba diver is ironing a dress shirt with an electric iron on an ironing board in a laundry room. \\
\rowcolor{tierFail!18}
gpt-5.1 & R2VR & iron & A hotel laundry worker is ironing a dress shirt with an iron on an ironing board in a laundry room. & A hotel bellhop is ironing a dress shirt with an iron on an ironing board in a laundry room. & A hairdresser is ironing a dress shirt with an iron on an ironing board in a laundry room. & A scuba diving instructor is ironing a dress shirt with an iron on an ironing board in a laundry room. \\
\rowcolor{tierFail!18}
sonnet-4.6 & RB & brush & A stable hand is brushing a horse with a grooming brush in a stable. & A jockey is brushing a horse with a grooming brush in a stable. & A veterinarian is brushing a horse with a grooming brush in a stable. & A chef is brushing a horse with a grooming brush in a stable. \\
\rowcolor{tierFail!18}
sonnet-4.6 & R2R & brush & A house painter is brushing a wall with a paintbrush in a house. & A construction worker is brushing a wall with a paintbrush in a house. & A plasterer is brushing a wall with a paintbrush in a house. & \textbf{A surgeon is brushing a wall with a paintbrush in a house.} \\
\rowcolor{tierFail!18}
sonnet-4.6 & R2VR & brush & A dog groomer is brushing a dog's coat with a grooming brush at the pet salon. & A veterinarian is brushing a dog's coat with a grooming brush at the pet salon. & A circus ringmaster is brushing a dog's coat with a grooming brush at the pet salon. & A chef is brushing a dog's coat with a grooming brush at the pet salon. \\
\rowcolor{tierFail!18}
gpt-5.1 & RB & carry & A warehouse worker is carrying a large cardboard box with both arms in a warehouse aisle. & A delivery driver is carrying a large cardboard box with both arms in a warehouse aisle. & A construction worker is carrying a large cardboard box with both arms in a warehouse aisle. & A ballet dancer is carrying a large cardboard box with both arms in a warehouse aisle. \\
\bottomrule
\end{tabularx}
\caption{Generation Examples}
\label{tab:v7-agent-paper-examples}
\vspace{2pt}
{\footnotesize
\colorbox{tierGold!35}{\strut\phantom{Gg}} GOLD\quad\colorbox{tierSilver!35}{\strut\phantom{Gg}} SILVER\quad\colorbox{tierBronze!35}{\strut\phantom{Gg}} BRONZE\quad\colorbox{tierFail!55}{\strut\phantom{Gg}} FAIL
\\
\textit{FAIL rows:} sentence highlighting is applied only when `scene\_verdict=GOOD`.
}
\end{table*}